\documentclass{article} 
\usepackage{iclr2026_conference,times}

\usepackage{hyperref}
\usepackage{url}
\usepackage{relsize}
\usepackage{graphicx}
\usepackage{mathtools}

\title{Near-Optimal Sample Complexity Bounds
for Constrained Average-Reward MDPs}

\author{Yukuan Wei \\
School of Mathematical Sciences\\
Fudan University\\
\texttt{22300180145@m.fudan.edu.cn} \\
\And
Xudong Li \\
School of Data Science \\
Fudan University \\
\texttt{lixudong@fudan.edu.cn} \\
\And
Lin F. Yang \\
Electrical and Computer Engineering / Computer Science\\
University of California, Los Angeles \\
\texttt{linyang@ee.ucla.edu} \\
}

\newcommand{\polylog}{\operatorname{polylog}}

\usepackage[utf8]{inputenc} 
\usepackage[T1]{fontenc}    
\usepackage{url}            
\usepackage{booktabs}       
\usepackage{amsfonts}       
\usepackage{xcolor}         
\usepackage{amsthm}

\usepackage{amsmath}
\usepackage{mathrsfs}
\usepackage{amssymb}
\usepackage{amssymb}
\usepackage{mathtools}
\usepackage{bm}
\usepackage{datetime}
\usepackage{tikz}
\usepackage{listings}
\usepackage{mdframed}
\usepackage{hyperref}
\hypersetup{
    unicode=false,          
    pdftoolbar=true,        
    pdfmenubar=true,        
    pdffitwindow=false,     
    pdfstartview={FitH},    
    pdftitle={XXXXX},    
    pdfauthor={XXX},     
    pdfsubject={Bandits, Reinforcement Learning},   
    pdfcreator={Creator},   
    pdfproducer={Producer}, 
    pdfkeywords={bandits} {reinforcement learning} {policy gradient}, 
    pdfnewwindow=true,      
    colorlinks=true,       
    linkcolor=blue,          
    citecolor=blue,        
    filecolor=magenta,      
    urlcolor=cyan           
}
\usepackage{times}
\usepackage{nicefrac}
\usepackage{wrapfig}

\usepackage{xpatch}
\usepackage{thm-restate}
\usepackage{subfigure}
\definecolor{shadecolor}{gray}{0.90}
\declaretheoremstyle[
headfont=\normalfont\bfseries,
notefont=\mdseries, notebraces={(}{)},
bodyfont=\normalfont,
postheadspace=0.5em,
spaceabove=5pt,
mdframed={
  skipabove=3pt,
  skipbelow=3pt,
  hidealllines=true,
  backgroundcolor={shadecolor},
  innerleftmargin=2pt,
  innerrightmargin=2pt}
]{shaded}

\usepackage[algoruled, linesnumbered]{algorithm2e}
\usepackage[capitalize]{cleveref}

\declaretheorem[style=shaded]{theorem}

\newtheorem{lemma}[theorem]{Lemma}

\newtheorem{assumption}[theorem]{Assumption}

\crefname{alg}{Algorithm}{Algorithm}

\usepackage{dsfont}
\usepackage{colortbl}
\usepackage{pifont}
\usepackage{microtype} 
\usepackage{float}
\usepackage{xfrac} 
\usepackage{xspace}

\makeatletter
\renewcommand*{\@textcolor}[3]{%
  \protect\leavevmode
  \begingroup
    \color#1{#2}#3%
  \endgroup
}
\makeatother

\mdfdefinestyle{mdframedthmbox}{%
	leftmargin=.0\textwidth,
	rightmargin=.0\textwidth,%
	innertopmargin=0.75em,
	innerleftmargin=.5em,
	innerrightmargin=.5em,
}
\newenvironment{thmbox}
	{%
		\begin{mdframed}[style=mdframedthmbox]%
	}{%
		\end{mdframed}%
	}

\usepackage{wrapfig}

\newcommand{\E}{\mathbb{E}}

\newcommand{\cA}{\mathcal{A}}

\newcommand{\cR}{\mathcal{R}}
\newcommand{\cS}{\mathcal{S}}

\renewcommand{\epsilon}{\varepsilon}

\DeclareMathOperator*{\argmax}{arg\ max}

\newcommand{\Clim}{\mathop{\mathrm{C\text{-}lim}}}
\newcommand{\spannorm}[1]{\left\lVert #1 \right\rVert_{\mathrm{span}}}

\newcommand{\const}[1]{\rho_c^{#1}(s)}

\newcommand{\reward}[1]{\rho_r^{#1}(s)}

\newcommand{\piopt}{\pi^*}
\newcommand{\pihat}{\hat{\pi}}

\providecommand{\cS}{\mathcal S}
\providecommand{\cA}{\mathcal A}

\providecommand{\pihat}{\widehat\pi}
\providecommand{\piopt}{\pi^\star}
\providecommand{\spannorm}[1]{\left\|#1\right\|_{\mathrm{span}}}

\makeatletter
\@ifundefined{assumption}{
  \newtheorem{assumption}{Assumption}
}{}
\makeatother

\hypersetup{
  pdftitle={Near-Optimal Sample Complexity Bounds for Constrained Average-Reward MDPs},
  pdfauthor={Yukuan Wei, Xudong Li, Lin F. Yang},
  pdfkeywords={constrained average-reward MDPs, generative model, sample complexity, minimax lower bound}
}

\iclrfinalcopy

\begin{document}
\maketitle

\begin{abstract}
We study constrained average-reward Markov decision processes (CAMDPs) under a generative model. Both the benchmark and the output are external mixtures of stationary deterministic policies. We consider relaxed feasibility, which allows an $\epsilon$ constraint violation, and strict feasibility, which requires exact constraint satisfaction. Our model-based primal--dual algorithm solves perturbed empirical discounted MDPs and transfers their guarantees to the average-reward problem. Let $H$ be a uniform upper bound on the reward and constraint bias spans, $B$ a uniform transient-time bound, and $\zeta$ a positive lower bound on the Slater margin. A cross-signal concentration theorem controls precisely the finite family of empirical Lagrangian oracle policies used by the algorithm through perturbed action gaps, row decoupling, and an instance-dependent Bernstein hierarchy. Under a policy-wise discounted variance condition, the relaxed and strict sample complexities are $\widetilde O\bigl(SA(B+H)/\epsilon^2\bigr)$ and $\widetilde O\bigl(SA(B+H)/(\epsilon^2\zeta^2)\bigr)$, respectively. For weakly communicating CAMDPs, we construct a hard family of diameter $D$ with actual span $H=\Theta(D)$ and obtain the adaptive minimax lower bound $\widetilde\Omega\bigl(SAH/(\epsilon^2\zeta^2)\bigr)$. A separate transient hard family gives $\widetilde\Omega\bigl(SAB/(\epsilon^2\zeta^2)\bigr)$ under balanced sampling. Since both hard subfamilies belong to the general CAMDP class, their worst-case lower bounds combine to $\widetilde\Omega\bigl(SA(B+H)/(\epsilon^2\zeta^2)\bigr)$ for balanced algorithms, matching the strict-feasibility upper rate up to logarithmic factors.
\end{abstract}

\section{Introduction}
\label{sec:intro}

Reinforcement learning (RL)~\citep{sutton1998reinforcement} provides a powerful framework for sequential decision-making under uncertainty, with successes in domains such as game playing~\citep{mnih2015human,silver2016mastering}, robotic control~\citep{tan2018sim,zeng2020tossingbot}, clinical decision-making~\citep{schaefer2005modeling}, and aligning large language models with human preferences~\citep{ouyang2022training}. Most classical RL algorithms optimize a single reward signal without additional constraints. Yet in many high-stakes applications, agents must also be safe, fair, and resource-aware, in addition to being efficient in practice. This leads to \emph{constrained Markov decision processes} (CMDPs)~\citep{altman1999constrained}, which maximize expected reward subject to an auxiliary cost constraint. A representative example arises in wireless sensor networks~\citep{buratti2009overview,julian2002qos}, balancing high throughput with average power constraints.

A growing body of work investigates constrained reinforcement learning in unknown environments~\citep{efroni2020exploration,zheng2020constrained,qiu2020upper,brantley2020constrained,kalagarla2020sample,yu2021provably,ding2021provably,gattami2021reinforcement,miryoosefi2022simple}. These efforts focus on the online learning setting, aiming to minimize both regret and constraint violation under exploration, estimation, and policy optimization challenges in finite-state, finite-action CMDPs. In contrast, recent work~\citep{hasan2021model,wei2021provably,bai2021achieving,vaswani_near-optimal_2022} studies the \emph{generative model}~\citep{kearns1999finite,kakade2003sample,agarwal2020model,sidford_near-optimal_2018,yang2019sample}, i.e., a simulator that provides sample transitions and rewards for any queried state-action pair, removing the need for exploration.

Most prior work centers on finite-horizon or discounted MDPs, where either the horizon is fixed to $T$ steps or future rewards are geometrically discounted by $\gamma^t$. While analytically convenient, these formulations can limit long-term performance: finite-horizon methods impose a cutoff, while discounting attenuates future rewards. To address this, the \emph{average-reward MDP} (AMDP) framework~\citep{puterman2014markov} seeks to maximize steady-state long-run average reward.

Although planning in AMDPs is relatively well-understood~\citep{altman1999constrained,borkar2005actor,borkar2014risk}, sample complexity for learning $\epsilon$-optimal policies was challenging due to the lack of episode resets and the need to reason about long-term behavior without discounting. Recent advances in the generative model setting establish near-optimal bounds depending on the \emph{optimal bias span} $H$ and the \emph{mixing or transient time} $B$, yielding rates $\tilde{\Theta}\!\left(\frac{SA(B + H)}{\epsilon^2}\right)$~\citep{zurek2024span}, where $S$ is the number of states and $A$ the number of actions.

Despite this progress for unconstrained AMDPs, the constrained variant—\emph{constrained average-reward MDPs} (CAMDPs)—remains poorly understood. In CAMDPs, the agent must maximize steady-state average reward while satisfying an average constraint on cost, risk, or resource usage, capturing long-term fairness, sustainability, and safe deployment. While discounted CMDPs have seen progress in both relaxed and strict feasibility regimes~\citep{vaswani_near-optimal_2022}, there are still no known sample complexity bounds for learning in CAMDPs, and the fundamental statistical limits in relaxed or strict feasibility settings remain open.

This gap motivates our work. We initiate the study of the sample complexity of learning $\epsilon$-optimal policies in CAMDPs under the generative model. We develop a model-based primal-dual algorithm handling both \emph{relaxed feasibility}, where the returned policy may violate the constraint by at most $\epsilon$, and \emph{strict feasibility}, where the policy must satisfy it exactly. We establish matching near-optimal upper and lower bounds in terms of the bias span, the transient time, and the \emph{Slater constant} $\zeta$. While relaxed and strict feasibility have been studied in discounted CMDPs~\citep{vaswani_near-optimal_2022}, our work provides the first sample complexity characterization for CAMDPs in the average-reward setting. Below, we summarize our contributions in more detail.

\paragraph{Contributions.}
We present the first near-optimal sample-complexity bounds for learning in CAMDPs with access to a generative model:\\
$\bullet$ We design a model-based primal--dual algorithm that returns an $\epsilon$-optimal external mixture of stationary deterministic policies under both relaxed and strict feasibility. Our method solves a sequence of perturbed empirical discounted MDPs with Lagrangian rewards and transfers their guarantees to the original average-reward problem. We further establish a cross-signal concentration result for the finite family of empirical Lagrangian oracle policies generated by the algorithm, avoiding a uniform concentration requirement over the entire policy class.\\
$\bullet$ In the relaxed-feasibility setting, we prove that our algorithm requires at most
$\widetilde O\left(\frac{SA(B+H)}{\epsilon^2}\right)$ samples, where $S$ and $A$ are the numbers of states and actions, $H$ is a uniform upper bound on the reward and constraint bias spans of stationary deterministic policies, and $B$ is a uniform transient-time bound. Notably, the relaxed-feasibility rate does not depend polynomially on the Slater margin.\\
$\bullet$ In the strict-feasibility setting, the sample complexity increases to
$\widetilde O\left(\frac{SA(B+H)}{\epsilon^2\zeta^2}\right)$, where $\zeta$ is a positive lower bound on the Slater margin characterizing the size of the feasible region. We prove an adaptive minimax lower bound
$\widetilde\Omega\left(\frac{SAH}{\epsilon^2\zeta^2}\right)$ for a weakly communicating family with actual uniform span $H=\Theta(D)$, as well as a transient-time lower bound
$\widetilde\Omega\left(\frac{SAB}{\epsilon^2\zeta^2}\right)$ under the balanced generative-model protocol. These lower bounds show that the additional $\zeta^{-2}$ dependence required for strict feasibility is unavoidable and establish a provable separation between the relaxed and strict regimes.

Together, our results provide the first near minimax-optimal sample-complexity characterization for constrained average-reward reinforcement learning and reveal how the bias span, transient behavior, and feasible-region margin jointly determine the statistical difficulty of learning under long-run constraints.

\subsection{Related Work}
There is a large body of research on the sample complexity of learning in \emph{unconstrained} Markov decision processes (MDPs); see the monograph by~\citet{agarwal2019reinforcement} for a comprehensive overview. In parallel, substantial progress has been made in \emph{constrained} reinforcement learning under unknown dynamics~\citep{efroni2020exploration,zheng2020constrained,qiu2020upper,brantley2020constrained,kalagarla2020sample,yu2021provably,ding2021provably,gattami2021reinforcement,miryoosefi2022simple}, particularly in finite-horizon settings. Another line of work addresses \emph{discounted} constrained MDPs (CMDPs) with access to a generative model~\citep{hasan2021model,wei2021provably,bai2021achieving,vaswani_near-optimal_2022}, yielding sample-efficient algorithms under both relaxed and strict constraint satisfaction.

In contrast, the average-reward setting is less explored. For unconstrained average-reward MDPs, \citet{zurek2024span} established nearly minimax-optimal bounds under a generative model, showing that $\widetilde{O}\left(SA H / \varepsilon^2\right)$ samples suffice for weakly communicating MDPs, where $H$ is the span of the optimal bias function. They further introduced a transient time parameter $B$ to handle general multichain MDPs, proving a matching bound of $\widetilde{O}\left(SA(B + H)/\varepsilon^2\right)$. However, their analysis does not incorporate constraints, and extending their framework to constrained average-reward MDPs (CAMDPs) remains open.

Among works on CMDPs, \citet{vaswani_near-optimal_2022} provided the first minimax-optimal sample complexity bounds for the \emph{discounted} setting via dual linear programming. Yet their techniques do not extend to average-reward problems, where key properties like Bellman contraction no longer hold. In a separate effort, \citet{bai2024learning} studied CAMDPs in an online model-free setting with general policy classes, establishing sublinear regret for constraint violation and the duality gap. Their results, however, focus on asymptotic behavior and do not provide near-optimal finite-sample guarantees under a generative model.

To our knowledge, this work is the first to give matching strict-feasibility rates for CAMDPs at the $B+H$ scale within the balanced generative-model protocol, together with an adaptive $H$-scale lower bound for weakly communicating CAMDPs.

\section{Problem Formulation}
\label{sec:problem}

A CAMDP is $M=\langle\cS,\cA,P,r,c,b,s\rangle$, where $P:\cS\times\cA\to\Delta_{\cS}$, $r,c:\cS\times\cA\to[0,1]$, $b\in[0,1]$, and $s\in\Delta_{\cS}$. Let $S=|\cS|$, $A=|\cA|$, and let $\Pi_{\mathrm{det}}$ be the stationary deterministic policy class. We optimize over external mixtures $\Pi_{\mathrm{mix}}:=\Delta(\Pi_{\mathrm{det}})$: one deterministic policy is drawn at the beginning of a trajectory and followed forever.

For $q\in\{r,c\}$ and $\pi\in\Pi_{\mathrm{det}}$, define
\begin{align}
\rho_q^\pi(x)
&:=\lim_{T\to\infty}\frac1T\E_x^\pi\!\left[\sum_{t=0}^{T-1}q(S_t,A_t)\right],
\nonumber\\
h_q^\pi(x)
&:=\Clim_{T\to\infty}\E_x^\pi\!\left[\sum_{t=0}^{T-1}\bigl(q(S_t,A_t)-\rho_q^\pi(S_t)\bigr)\right].
\label{eq:gain-bias}
\end{align}
The pair $(\rho_q^\pi,h_q^\pi)$ satisfies $\rho_q^\pi=P_\pi\rho_q^\pi$ and $\rho_q^\pi+h_q^\pi=q_\pi+P_\pi h_q^\pi$. Write $\rho_q^\pi(s)=\sum_xs(x)\rho_q^\pi(x)$ and extend all values linearly to external mixtures.

The benchmark is
\begin{align}
\piopt\in\argmax_{\pi\in\Pi_{\mathrm{mix}}}\reward{\pi}
\quad\mathrm{s.t.}\quad \const{\pi}\ge b.
\label{eq:true-CMDP}
\end{align}
An optimizer can be chosen with support size at most two. For each $\pi\in\Pi_{\mathrm{det}}$, let $\cR^\pi$ be its recurrent states and $T_{\cR^\pi}:=\inf\{t\ge0:S_t\in\cR^\pi\}$. We assume supplied structural bounds
\begin{align}
\spannorm{h_r^\pi}\le H,
\qquad
\spannorm{h_c^\pi}\le H,
\qquad
\forall\pi\in\Pi_{\mathrm{det}},
\label{eq:def-H}
\end{align}
and
\begin{align}
\max_{x\in\cS}\E_x^\pi[T_{\cR^\pi}]\le B,
\qquad
\forall\pi\in\Pi_{\mathrm{det}}.
\label{eq:def-B}
\end{align}
Let $\pi_c^*\in\argmax_{\pi\in\Pi_{\mathrm{det}}}\const{\pi}$. Feasibility gives $\const{\pi_c^*}\ge b$; for strict feasibility we are supplied with $\zeta>0$ satisfying
\begin{align}
\const{\pi_c^*}\ge b+\zeta.
\label{eq:def-zeta}
\end{align}
We work in the nontrivial regime $B+H\ge1$.

The learner knows $r,c$ and, for the upper bound, obtains $n$ independent transitions from every row $P(\cdot\mid x,a)$, so $N_{\mathrm{tot}}=SAn$. Relaxed feasibility requires $\reward{\pihat}\ge\reward{\piopt}-\epsilon$ and $\const{\pihat}\ge b-\epsilon$; strict feasibility replaces the latter by $\const{\pihat}\ge b$.

For $\gamma\in(0,1)$, define
\[
V_{q,\gamma}^\pi(x):=\E_x^\pi\!\left[\sum_{t\ge0}\gamma^tq(S_t,A_t)\right],
\qquad
V_{q,\gamma}^\pi(s):=\sum_xs(x)V_{q,\gamma}^\pi(x).
\]
Empirical values under $\widehat P$ are denoted by $\widehat V_{q,\gamma}^\pi$. We use $(1-\gamma)V_{q,\gamma}^\pi$ as the normalized discounted value without introducing a separate symbol. Let $\sigma_{q,\gamma}^\pi:=\sqrt{\operatorname{Var}_{P_\pi}(V_{q,\gamma}^\pi)}$ entrywise.

\begin{assumption}[Policy-wise discounted variance condition]
\label{assum:policywise-variance}
There is a universal $C_{\mathrm{var}}>0$ such that, whenever $B+H\le(1-\gamma)^{-1}$,
\begin{align}
\gamma\left\|(I-\gamma P_\pi)^{-1}\sigma_{q,\gamma}^\pi\right\|_\infty
\le C_{\mathrm{var}}\frac{\sqrt{B+H}}{1-\gamma}
\label{eq:variance-condition}
\end{align}
for every $\pi\in\Pi_{\mathrm{det}}$ and $q\in\{r,c\}$.
\end{assumption}

\section{Methodology}
\label{sec:method}

\begin{algorithm}[t]
\caption{Discounted Model-Based Algorithm for CAMDPs}
\label{alg:camdp}
\textbf{Input:} $n,\gamma,b',U,\omega,\varepsilon_1,T,\eta$, and $\lambda_0=0$.\\
Collect $n$ samples per $(x,a)$ and form $\widehat P$. Independently draw $Z(x,a)\sim\operatorname{Unif}[0,\omega]$ and set $r_p=r+Z$. Form $\Lambda=\{k\varepsilon_1:0\le k\le\lfloor U/\varepsilon_1\rfloor\}\cup\{U\}$.\\
\textbf{for} $t=0,\ldots,T-1$ \textbf{do}\\
\quad Compute $\pihat_t\in\Pi_{\mathrm{det}}$ satisfying
$\widehat V_{r_p+\lambda_tc,\gamma}^{\pihat_t}=\widehat V_{r_p+\lambda_tc,\gamma}^{*}$ componentwise.\\
\quad Set $\lambda_{t+1}=\mathcal R_\Lambda[\mathbb P_{[0,U]}[\lambda_t-\eta((1-\gamma)\widehat V_{c,\gamma}^{\pihat_t}(s)-b')]]$.\\
\textbf{end for}\\
\textbf{Output:} the external mixture $\pihat=T^{-1}\sum_{t=0}^{T-1}\pihat_t$.
\end{algorithm}

The empirical reference CMDP is
\begin{align}
\widehat\pi_\gamma^*\in\argmax_{\pi\in\Pi_{\mathrm{mix}}}(1-\gamma)\widehat V_{r_p,\gamma}^{\pi}(s)
\quad\mathrm{s.t.}\quad (1-\gamma)\widehat V_{c,\gamma}^{\pi}(s)\ge b'.
\label{eq:change-CAMDP}
\end{align}
Strong duality holds. Let $\lambda^*$ be an optimal dual multiplier. The statewise oracle also satisfies
\begin{align}
\pihat_t\in\argmax_{\pi\in\Pi_{\mathrm{det}}}
\left\{(1-\gamma)\widehat V_{r_p,\gamma}^{\pi}(s)+\lambda_t(1-\gamma)\widehat V_{c,\gamma}^{\pi}(s)\right\},
\label{eq:primal-update}
\end{align}
while the dual update is
\begin{align}
\lambda_{t+1}=\mathcal R_\Lambda\!\left[\mathbb P_{[0,U]}\!\left[\lambda_t-\eta\bigl((1-\gamma)\widehat V_{c,\gamma}^{\pihat_t}(s)-b'\bigr)\right]\right].
\label{eq:dual-update}
\end{align}
For every $\lambda\in\Lambda$, let $\pihat_\lambda$ be the lexicographically selected statewise optimizer satisfying
\begin{align}
\widehat V_{r_p+\lambda c,\gamma}^{\pihat_\lambda}=\widehat V_{r_p+\lambda c,\gamma}^{*}.
\label{eq:oracle-policy}
\end{align}
The rounding map selects the smaller grid point in a tie.

\begin{restatable}[Cross-signal concentration for Lagrangian oracle policies]{theorem}{oracleconcentration}
\label{thm:cross-discount-conc}
Suppose Assumption~\ref{assum:policywise-variance} holds, $B+H\le(1-\gamma)^{-1}$, and $\alpha,\delta\in(0,1)$. If
\[
n\ge C\left(\frac{B+H}{\alpha^2}+\frac1{\alpha(1-\gamma)}\right)
\polylog\!\left(
S,A,|\Lambda|,\frac1\delta,\frac1\alpha,\frac1\omega,
\frac1{1-\gamma},1+U,B+H
\right),
\]
then, with probability at least $1-\delta$, simultaneously for every $\lambda\in\Lambda$ and $q\in\{r,c\}$,
\[
(1-\gamma)\left\|V_{q,\gamma}^{\pihat_\lambda}-\widehat V_{q,\gamma}^{\pihat_\lambda}\right\|_\infty\le\alpha.
\]
The same rate holds for any fixed $O(1)$ collection of deterministic policies independent of the samples and for external mixtures supported on covered policies.
\end{restatable}

The proof is in Appendix~\ref{app:concentration}. Reward perturbation creates a simultaneous action gap. For each transition row, a state--action absorbing construction puts the empirical oracle in a finite candidate set independent of that row. Conditional Bernstein inequalities then verify the hierarchy required to retain the instance-dependent propagated-variance term.

\begin{restatable}[Primal--dual guarantee]{theorem}{pdguarantees}
\label{thm:pd-guarantees}
Let $G_{b'}=\max\{|b'|,|1-b'|\}$, $d=U-\lambda^*$, and $m=\min\{1,d\}$. If $U>\lambda^*$, $\eta=U/(G_{b'}\sqrt T)$,
$T\ge4U^2G_{b'}^2/(\varepsilon_{\mathrm{opt}}^2m^2)$, and
$0<\varepsilon_1\le\min\{U,\varepsilon_{\mathrm{opt}}m/(3G_{b'}\sqrt T)\}$, then
\[
(1-\gamma)\widehat V_{r_p,\gamma}^{\pihat}(s)\ge(1-\gamma)\widehat V_{r_p,\gamma}^{\widehat\pi_\gamma^*}(s)-\varepsilon_{\mathrm{opt}},
\qquad
(1-\gamma)\widehat V_{c,\gamma}^{\pihat}(s)\ge b'-\varepsilon_{\mathrm{opt}}.
\]
The guarantee is deterministic conditional on $(\widehat P,Z)$.
\end{restatable}

\section{Upper Bound under Relaxed Feasibility}
\label{sec:ub-relaxed}

We solve a slightly relaxed empirical discounted CMDP so that the approximation, estimation, and optimization errors are absorbed by the allowed constraint violation. Set
\begin{align*}
1-\gamma&=\frac{\epsilon}{16(B+H)}, &
\alpha_{\rm rel}&=\frac{\epsilon}{16}, &
\omega&=\frac{\epsilon}{16},\\
b'&=b-\frac{\epsilon}{4}, &
\varepsilon_{\rm opt}&=\frac{\epsilon}{4}, &
U&=\frac{18}{\epsilon}.
\end{align*}
Let $G_{\rm rel}=\max\{|b'|,|1-b'|\}\le5/4$ and choose
\[
T=\left\lceil\frac{4U^2G_{\rm rel}^2}{\varepsilon_{\rm opt}^2}\right\rceil,
\qquad
\eta=\frac{U}{G_{\rm rel}\sqrt T},
\qquad
\varepsilon_1=\min\left\{U,\frac{\varepsilon_{\rm opt}}{3G_{\rm rel}\sqrt T}\right\}.
\]
Then $T=O(\epsilon^{-4})$, $\varepsilon_1=O(\epsilon^3)$, and $|\Lambda|=O(\epsilon^{-4})$, so the grid contributes only logarithmic factors.

\begin{restatable}[Relaxed feasibility]{theorem}{ubrelaxed}
\label{thm:ub-relaxed}
Fix $\epsilon,\delta\in(0,1)$ and suppose Assumption~\ref{assum:policywise-variance} holds. There is a universal $C>0$ such that, if
\[
n\ge C\frac{B+H}{\epsilon^2}\polylog\!\left(S,A,\frac1\delta,\frac1\epsilon,\frac1\omega,\frac1{1-\gamma},|\Lambda|,1+U,B+H\right),
\]
then Algorithm~\ref{alg:camdp} returns an external mixture $\pihat$ satisfying $\rho_r^{\pihat}(s)\ge\rho_r^{\piopt}(s)-\epsilon$ and $\rho_c^{\pihat}(s)\ge b-\epsilon$ with probability at least $1-\delta$. Consequently,
\[
N_{\rm tot}=\widetilde O\!\left(\frac{SA(B+H)}{\epsilon^2}\right).
\]
\end{restatable}

\begin{proof}[Proof sketch]
Let $\kappa_\gamma=(1-\gamma)H\le\epsilon/16=\alpha_{\rm rel}$. Lemma~\ref{lemma:average-discount-bridge} gives $|(1-\gamma)V_{q,\gamma}^{\pi}(s)-\rho_q^\pi(s)|\le\kappa_\gamma$ for $q\in\{r,c\}$. Apply Theorem~\ref{thm:cross-discount-conc} with $\alpha=\alpha_{\rm rel}$ to the at most two deterministic policies in the support of $\piopt$, the fixed policy $\pi_c^*$, and all oracle policies $\pihat_\lambda$. By mixture linearity, the same event covers $\piopt$ and $\pihat$.

Since $\rho_c^{\pi_c^*}(s)\ge b$,
\[
(1-\gamma)\widehat V_{c,\gamma}^{\pi_c^*}(s)-b'
\ge \frac\epsilon4-\kappa_\gamma-\alpha_{\rm rel}
\ge\frac\epsilon8.
\]
Lemma~\ref{lemma:dual-bound} yields $\lambda^*\le9/\epsilon$, and hence $U-\lambda^*\ge1$. Moreover,
\[
(1-\gamma)\widehat V_{c,\gamma}^{\piopt}(s)
\ge\rho_c^{\piopt}(s)-\kappa_\gamma-\alpha_{\rm rel}
\ge b-\frac\epsilon8\ge b',
\]
so $\piopt$ is feasible for the empirical reference problem. Theorem~\ref{thm:pd-guarantees} therefore gives
\begin{align}
(1-\gamma)\widehat V_{r_p,\gamma}^{\pihat}(s)
&\ge(1-\gamma)\widehat V_{r_p,\gamma}^{\widehat\pi_\gamma^*}(s)-\varepsilon_{\rm opt},
\label{eq:rel-pd-r}\\
(1-\gamma)\widehat V_{c,\gamma}^{\pihat}(s)
&\ge b'-\varepsilon_{\rm opt}.
\label{eq:rel-pd-c}
\end{align}
The constraint transfers as
\[
\rho_c^{\pihat}(s)
\ge b'-\varepsilon_{\rm opt}-\alpha_{\rm rel}-\kappa_\gamma
=b-\frac{5\epsilon}{8}\ge b-\epsilon.
\]

For reward, decompose
\begin{align}
\rho_r^{\piopt}(s)-\rho_r^{\pihat}(s)
={}&\underbrace{\rho_r^{\piopt}(s)-(1-\gamma)V_{r,\gamma}^{\piopt}(s)}_{\rm bridge}
+\underbrace{(1-\gamma)(V_{r,\gamma}^{\piopt}-\widehat V_{r,\gamma}^{\piopt})(s)}_{\rm fixed\ estimation}
\nonumber\\
&+\underbrace{(1-\gamma)(\widehat V_{r,\gamma}^{\piopt}-\widehat V_{r_p,\gamma}^{\piopt})(s)}_{\le0}
+\underbrace{(1-\gamma)(\widehat V_{r_p,\gamma}^{\piopt}-\widehat V_{r_p,\gamma}^{\widehat\pi_\gamma^*})(s)}_{\le0}
\nonumber\\
&+\underbrace{(1-\gamma)(\widehat V_{r_p,\gamma}^{\widehat\pi_\gamma^*}-\widehat V_{r_p,\gamma}^{\pihat})(s)}_{\rm optimization}
+\underbrace{(1-\gamma)(\widehat V_{r_p,\gamma}^{\pihat}-\widehat V_{r,\gamma}^{\pihat})(s)}_{\rm perturbation}
\nonumber\\
&+\underbrace{(1-\gamma)(\widehat V_{r,\gamma}^{\pihat}-V_{r,\gamma}^{\pihat})(s)}_{\rm oracle\ estimation}
+\underbrace{(1-\gamma)V_{r,\gamma}^{\pihat}(s)-\rho_r^{\pihat}(s)}_{\rm bridge}.
\label{eq:relaxed-body-decomposition}
\end{align}
The bridge terms are at most $\kappa_\gamma$, the two estimation terms at most $\alpha_{\rm rel}$, the optimization term at most $\varepsilon_{\rm opt}$ by~\cref{eq:rel-pd-r}, and the perturbation term at most $\omega$. Thus
\[
\rho_r^{\piopt}(s)-\rho_r^{\pihat}(s)
\le2\kappa_\gamma+2\alpha_{\rm rel}+\omega+\varepsilon_{\rm opt}
\le\frac{9\epsilon}{16}<\epsilon.
\]
Finally, Theorem~\ref{thm:cross-discount-conc} requires
$ n\gtrsim (B+H)/\alpha_{\rm rel}^2+[\alpha_{\rm rel}(1-\gamma)]^{-1}$ up to logs. Both terms are $O((B+H)/\epsilon^2)$.
\end{proof}

\section{Upper Bound under Strict Feasibility}
\label{sec:ub-strict}

Strict feasibility is obtained by tightening the empirical constraint. Choose
\begin{align*}
1-\gamma&=\frac{\epsilon\zeta}{128(B+H)}, &
\alpha_{\rm str}&=\frac{\epsilon\zeta}{128}, &
\omega&=\frac\epsilon{16},\\
\Delta&=\frac{\epsilon\zeta}{32}, &
b'&=b+\Delta, &
\varepsilon_{\rm opt}&=\frac{\epsilon\zeta}{128},
\qquad U=\frac8\zeta.
\end{align*}
Assume $b+\zeta\le1$. With $G_{\rm str}=\max\{|b'|,|1-b'|\}\le1$, choose
\[
T=\left\lceil\frac{4U^2G_{\rm str}^2}{\varepsilon_{\rm opt}^2}\right\rceil,
\quad
\eta=\frac{U}{G_{\rm str}\sqrt T},
\quad
\varepsilon_1=\min\left\{U,\frac{\varepsilon_{\rm opt}}{3G_{\rm str}\sqrt T}\right\}.
\]
Then $T=O((\epsilon^2\zeta^4)^{-1})$, $\varepsilon_1=O(\epsilon^2\zeta^3)$, and $|\Lambda|=O((\epsilon^2\zeta^4)^{-1})$.

\begin{restatable}[Strict feasibility]{theorem}{ubstrict}
\label{thm:ub-strict}
Fix $\epsilon,\delta\in(0,1)$, suppose $b+\zeta\le1$, and suppose Assumption~\ref{assum:policywise-variance} holds. There is a universal $C>0$ such that, if
\[
n\ge C\frac{B+H}{\epsilon^2\zeta^2}\polylog\!\left(S,A,\frac1\delta,\frac1\epsilon,\frac1\zeta,\frac1\omega,\frac1{1-\gamma},|\Lambda|,1+U,B+H\right),
\]
then Algorithm~\ref{alg:camdp} returns an external mixture $\pihat$ satisfying $\rho_r^{\pihat}(s)\ge\rho_r^{\piopt}(s)-\epsilon$ and $\rho_c^{\pihat}(s)\ge b$ with probability at least $1-\delta$. Consequently,
\[
N_{\rm tot}=\widetilde O\!\left(\frac{SA(B+H)}{\epsilon^2\zeta^2}\right).
\]
\end{restatable}

\begin{proof}[Proof sketch]
Let $\kappa_\gamma=(1-\gamma)H\le\alpha_{\rm str}$ and apply Theorem~\ref{thm:cross-discount-conc} to the support of $\piopt$, $\pi_c^*$, and all oracle policies. Since $\rho_c^{\pi_c^*}(s)\ge b+\zeta$,
\[
(1-\gamma)\widehat V_{c,\gamma}^{\pi_c^*}(s)-b'
\ge\zeta-\Delta-\kappa_\gamma-\alpha_{\rm str}\ge\frac\zeta2.
\]
Hence $\lambda^*\le3/\zeta$ and $U-\lambda^*\ge1$. Theorem~\ref{thm:pd-guarantees} gives
\begin{align}
(1-\gamma)\widehat V_{r_p,\gamma}^{\pihat}(s)
&\ge(1-\gamma)\widehat V_{r_p,\gamma}^{\widehat\pi_\gamma^*}(s)-\varepsilon_{\rm opt},
\label{eq:str-pd-r}\\
(1-\gamma)\widehat V_{c,\gamma}^{\pihat}(s)
&\ge b+\Delta-\varepsilon_{\rm opt}.
\label{eq:str-pd-c}
\end{align}
Therefore
\[
\rho_c^{\pihat}(s)
\ge b+\Delta-\varepsilon_{\rm opt}-\alpha_{\rm str}-\kappa_\gamma\ge b.
\]

To control the reward loss caused by tightening, define $b_0=b-\kappa_\gamma-\alpha_{\rm str}$ and let $\widetilde\pi_\gamma^*$ be the empirical optimizer at threshold $b_0$. The policy $\piopt$ is feasible at $b_0$, and
\[
d_b:=b'-b_0=\Delta+\kappa_\gamma+\alpha_{\rm str}\le\frac{6\epsilon\zeta}{128}.
\]
Lemma~\ref{lemma:sensitivity} yields
\begin{align}
(1-\gamma)\widehat V_{r_p,\gamma}^{\widetilde\pi_\gamma^*}(s)
-(1-\gamma)\widehat V_{r_p,\gamma}^{\widehat\pi_\gamma^*}(s)
\le d_b\lambda^*\le\frac{18\epsilon}{128}.
\label{eq:strict-sensitivity-body}
\end{align}
Now decompose
\begin{align}
\rho_r^{\piopt}(s)-\rho_r^{\pihat}(s)
={}&\underbrace{\rho_r^{\piopt}(s)-(1-\gamma)V_{r,\gamma}^{\piopt}(s)}_{\rm bridge}
+\underbrace{(1-\gamma)(V_{r,\gamma}^{\piopt}-\widehat V_{r,\gamma}^{\piopt})(s)}_{\rm fixed\ estimation}
\nonumber\\
&+\underbrace{(1-\gamma)(\widehat V_{r,\gamma}^{\piopt}-\widehat V_{r_p,\gamma}^{\piopt})(s)}_{\le0}
+\underbrace{(1-\gamma)(\widehat V_{r_p,\gamma}^{\piopt}-\widehat V_{r_p,\gamma}^{\widetilde\pi_\gamma^*})(s)}_{\le0}
\nonumber\\
&+\underbrace{(1-\gamma)(\widehat V_{r_p,\gamma}^{\widetilde\pi_\gamma^*}-\widehat V_{r_p,\gamma}^{\widehat\pi_\gamma^*})(s)}_{\rm sensitivity}
+\underbrace{(1-\gamma)(\widehat V_{r_p,\gamma}^{\widehat\pi_\gamma^*}-\widehat V_{r_p,\gamma}^{\pihat})(s)}_{\rm optimization}
\nonumber\\
&+\underbrace{(1-\gamma)(\widehat V_{r_p,\gamma}^{\pihat}-\widehat V_{r,\gamma}^{\pihat})(s)}_{\rm perturbation}
+\underbrace{(1-\gamma)(\widehat V_{r,\gamma}^{\pihat}-V_{r,\gamma}^{\pihat})(s)}_{\rm oracle\ estimation}
\nonumber\\
&+\underbrace{(1-\gamma)V_{r,\gamma}^{\pihat}(s)-\rho_r^{\pihat}(s)}_{\rm bridge}.
\label{eq:strict-body-decomposition}
\end{align}
Using the bridge, concentration, perturbation, sensitivity, and primal--dual bounds,
\[
\rho_r^{\piopt}(s)-\rho_r^{\pihat}(s)
\le2\kappa_\gamma+2\alpha_{\rm str}+\omega+\varepsilon_{\rm opt}+d_b\lambda^*
<\epsilon.
\]
Finally, the two leading terms in Theorem~\ref{thm:cross-discount-conc} are both $O((B+H)/(\epsilon^2\zeta^2))$.
\end{proof}

\section{Lower Bound for Weakly Communicating CAMDPs}
\label{sec:WCLB}

\begin{restatable}[Lower bound for weakly communicating CAMDPs]{theorem}{lbcomm}
\label{thm:wcmdp}
There are universal constants $c_0,c_1,c_2,c_3>0$ such that, for sufficiently large $S,A$ with $A\ge3$, $0<\epsilon\le c_0$, $0<\zeta\le c_1$, and $D\ge c_2(\lceil\log_{A-1}S\rceil+1)$, there is a family of weakly communicating CAMDPs with Slater margin exactly $\zeta$, diameter and uniform transient parameter at most $c_3D$, and actual uniform bias-span parameter $H\in[c_3^{-1}D,c_3D]$. Any possibly adaptive generative-model algorithm that returns an $\epsilon/48$-optimal strictly feasible external mixture with probability at least $3/4$ on every instance must use
\[
N_{\rm tot}=\widetilde\Omega\!\left(\frac{SAD}{\epsilon^2\zeta^2}\right)
=\widetilde\Omega\!\left(\frac{SAH}{\epsilon^2\zeta^2}\right)
\]
expected transition samples on some instance.
\end{restatable}

\begin{figure}[t]
\centering
\begin{minipage}[t]{0.53\textwidth}
\centering
\IfFileExists{wcmdp.png}{\includegraphics[width=\textwidth]{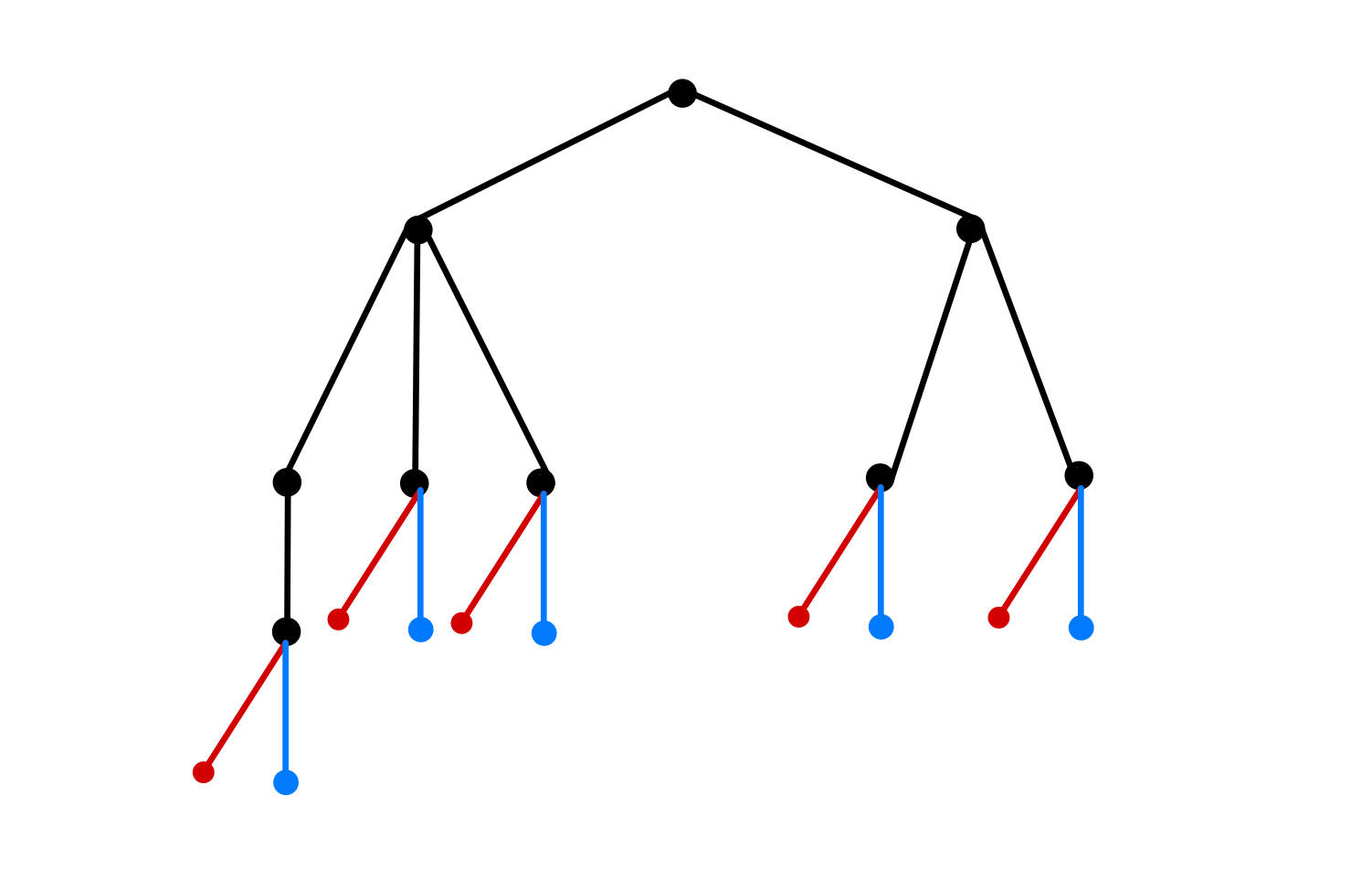}}{\fbox{\parbox[c][3.0cm][c]{0.92\textwidth}{\centering Place \texttt{wcmdp.png} here.}}}
\end{minipage}\hfill
\begin{minipage}[t]{0.41\textwidth}
\centering
\IfFileExists{Component.png}{\includegraphics[width=\textwidth]{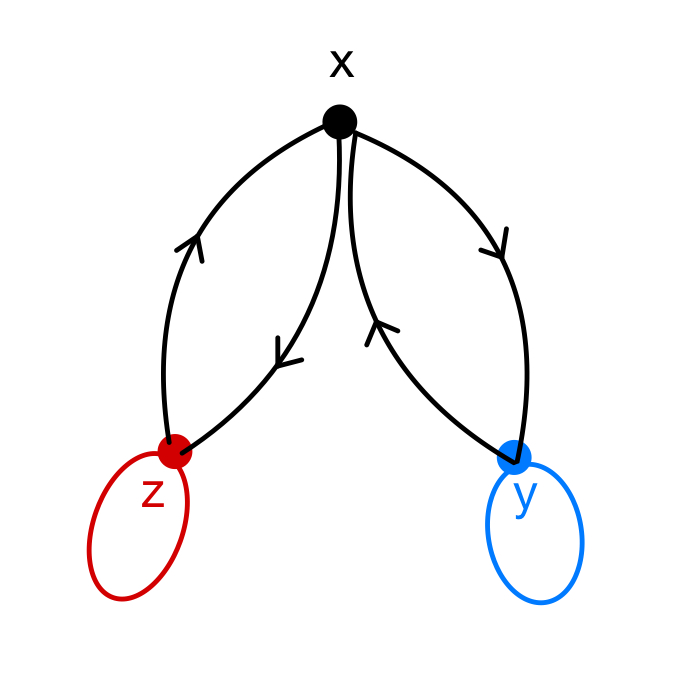}}{\fbox{\parbox[c][3.0cm][c]{0.90\textwidth}{\centering Place \texttt{Component.png} here.}}}
\end{minipage}
\caption{Left: a hard weakly communicating CAMDP obtained by embedding slow components at the leaves of a routing tree. Right: a primitive component used in the construction.}
\label{fig:weak-hard-family}
\end{figure}

\begin{proof}[Proof sketch]
Following the structure of the original construction, set $A'=A-1$, $D'=D/8$, and $K=\Theta(S)$. A primitive three-state component $(x_k,y_k,z_k)$ is attached to each leaf of an $A'$-ary routing tree; see Figure~\ref{fig:weak-hard-family}. The tree has depth $O(\log_{A'}S)$, while each slow component has holding time $\Theta(D')$, so the diameter is at most $D$.

The reference model $M_0$ and the perturbed models $M_{k,l}$ differ only at one component--action row. The constraint rewards amplify a transition perturbation of order $\epsilon\zeta/D$: every $\epsilon/48$-optimal strictly feasible policy must put normalized occupancy above $2/3$ on the baseline action in $M_0$, but below $2/3$ in $M_{k,l}$. The distinguished-row KL divergence is $O(\bigl(\epsilon^2\zeta^2\bigr)/D)$. A Fano/change-of-measure argument over $\Theta(SA)$ possible rows therefore gives
\[
N_{\rm tot}=\widetilde\Omega\!\left(\frac{SAD}{\epsilon^2\zeta^2}\right).
\]
For the constructed family, the component Poisson equations give a bias difference of order $D$ and the hitting-time representation gives the matching upper bound, so the actual uniform parameter satisfies $H=\Theta(D)$. Hence the same lower bound is $\widetilde\Omega(SAH/(\epsilon^2\zeta^2))$. Appendix~\ref{app:proofs-wclb} contains the complete construction and testing argument.
\end{proof}

\section{Lower Bound for General CAMDPs}
\label{sec:LBG}

\begin{restatable}[Lower bound for general CAMDPs]{theorem}{lbgen}
\label{thm:general-lower-bound}
There are universal constants $c_1,c_2>0$ such that, for sufficiently large $S,A$, $0<\epsilon\le c_1$, $0<\zeta\le c_2$, and structural scales $B,H\ge1$, any balanced generative-model algorithm that returns an $\epsilon/48$-optimal strictly feasible policy with probability at least $3/4$ uniformly over the general CAMDP class must use
\[
N_{\rm tot}=\widetilde\Omega\!\left(\frac{SA(B+H)}{\epsilon^2\zeta^2}\right)
\]
transition samples in the worst case.
\end{restatable}

\begin{figure}[t]
\centering
\IfFileExists{gmdp.png}{\includegraphics[width=0.72\textwidth]{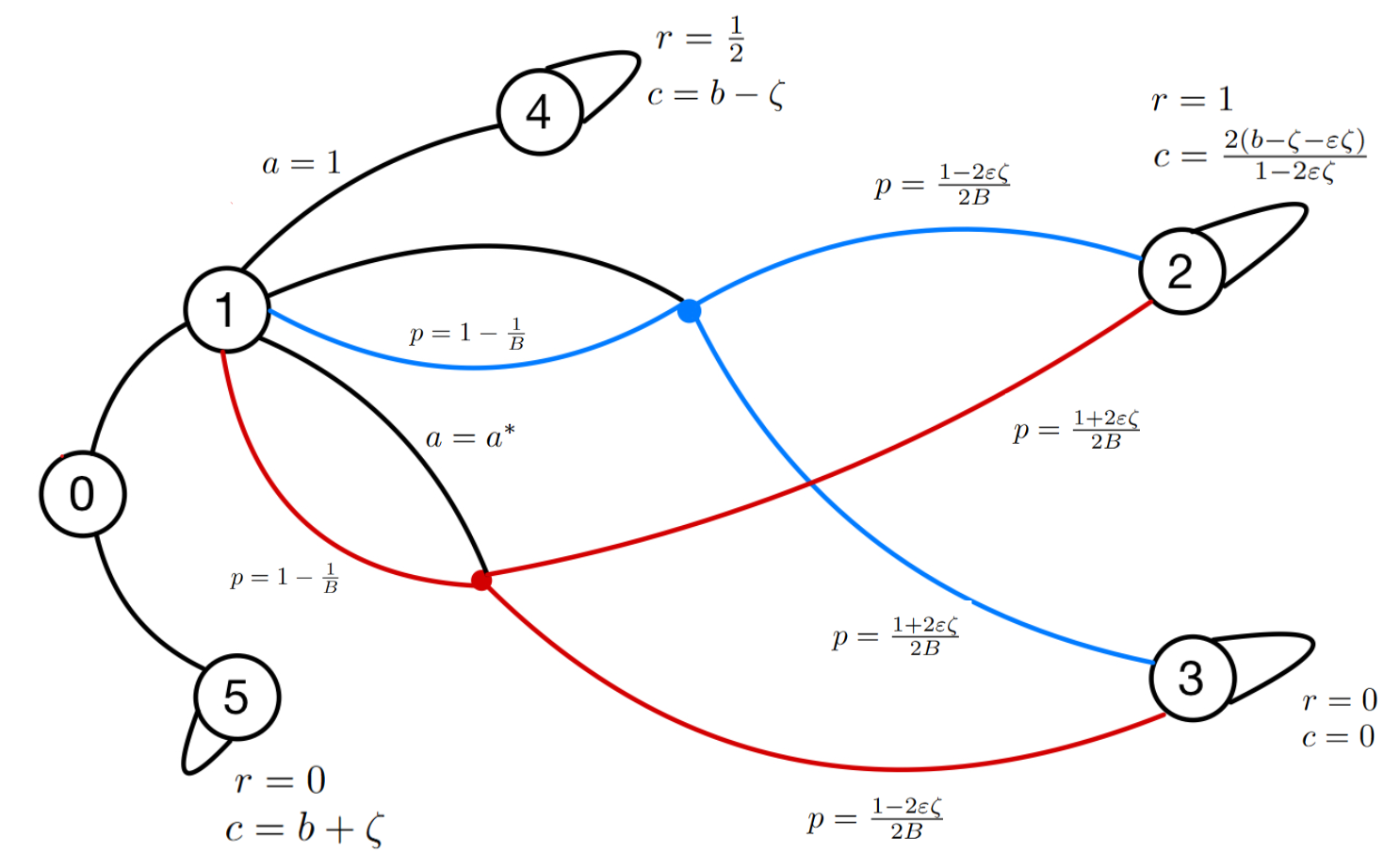}}{\fbox{\parbox[c][4.0cm][c]{0.68\textwidth}{\centering Place \texttt{gmdp.png} here.}}}
\caption{A component MDP used in the transient hard family.}
\label{fig:gmdp}
\end{figure}

\begin{proof}[Proof sketch]
The transient construction in Appendix~\ref{app:proofs-lb} retains the original six-state occupancy calculation shown in Figure~\ref{fig:gmdp}. The base and perturbed instances differ in one row; four linear programs give a strict $2/3$ occupancy separation, and the row divergence is $O(\epsilon^2\zeta^2/B)$. Thus balanced sampling requires
\[
\widetilde\Omega\!\left(\frac{SAB}{\epsilon^2\zeta^2}\right)
\]
samples. Theorem~\ref{thm:wcmdp} applies a fortiori to balanced algorithms and gives the independent $H$-scale lower bound. Since both hard subfamilies are contained in the general CAMDP class, the worst-case minimax lower bound is their maximum; using $\max\{B,H\}\ge(B+H)/2$,
\[
\max\left\{\widetilde\Omega\!\left(\frac{SAB}{\epsilon^2\zeta^2}\right),
\widetilde\Omega\!\left(\frac{SAH}{\epsilon^2\zeta^2}\right)\right\}
=\widetilde\Omega\!\left(\frac{SA(B+H)}{\epsilon^2\zeta^2}\right).
\]
This is the standard worst-case combination of the $B$- and $H$-hard subfamilies.
\end{proof}

\section{Conclusion}
\label{conclusion}
Under the policy-wise discounted variance condition, the relaxed and strict upper bounds are $\widetilde O(SA(B+H)/\epsilon^2)$ and $\widetilde O(SA(B+H)/(\epsilon^2\zeta^2))$. The data-dependent concentration theorem is restricted to the finite Lagrangian oracle family actually generated by the algorithm. The weakly communicating construction gives an adaptive minimax $H$-scale lower bound, while the transient construction gives the $B$-scale lower bound under balanced sampling. Combining the two hard subfamilies yields the balanced worst-case lower bound $\widetilde\Omega(SA(B+H)/(\epsilon^2\zeta^2))$, matching the strict-feasibility upper rate up to logarithmic factors.

\section*{Acknowledgements}
The research of Xudong Li was supported in part by the National Key R\&D Program of China [Grant 2023YFA1009300] and the National Natural Science Foundation of China [Grants 12271107 and 12531014].
\newpage
\bibliography{ref}
\bibliographystyle{iclr2026_conference}

\newpage
\appendix
\section{Proofs for the Primal--Dual Algorithm}
\label{app:proofs-pd}

\pdguarantees*

\begin{proof}
For notational convenience, define
\[
f(\pi)
:=
(1-\gamma)\widehat V_{r_p,\gamma}^{\pi}(s),
\qquad
g(\pi)
:=
(1-\gamma)\widehat V_{c,\gamma}^{\pi}(s),
\]
and
\[
f^*
:=
f(\widehat\pi_\gamma^*).
\]
For every comparator \(\lambda\in[0,U]\), define
\begin{align}
R^d(\lambda,T)
:=
\sum_{t=0}^{T-1}
(\lambda_t-\lambda)
\left(
g(\pihat_t)-b'
\right).
\label{eq:pd-regret}
\end{align}

By the scalar best-response property, for every
\(\pi\in\Pi_{\mathrm{mix}}\),
\[
f(\pihat_t)+\lambda_tg(\pihat_t)
\geq
f(\pi)+\lambda_tg(\pi).
\]
Taking \(\pi=\widehat\pi_\gamma^*\) gives
\[
f^*-f(\pihat_t)
\leq
\lambda_t
\left(
g(\pihat_t)-g(\widehat\pi_\gamma^*)
\right).
\]
Since \(\widehat\pi_\gamma^*\) is feasible,
\[
g(\widehat\pi_\gamma^*)
\geq
b',
\]
and therefore
\begin{align}
f^*-f(\pihat_t)
\leq
\lambda_t
\left(
g(\pihat_t)-b'
\right).
\label{eq:pd-bound}
\end{align}

Adding
\[
\lambda
\left(
b'-g(\pihat_t)
\right)
\]
to both sides, summing over \(t\), and dividing by \(T\) gives
\begin{align}
&
\frac1T
\sum_{t=0}^{T-1}
\left(
f^*-f(\pihat_t)
\right)
+
\frac{\lambda}{T}
\sum_{t=0}^{T-1}
\left(
b'-g(\pihat_t)
\right)
\nonumber
\\
&\hspace{35mm}
\leq
\frac{
R^d(\lambda,T)
}{
T
}.
\label{eq:pd-before-mixture}
\end{align}

By external-mixture linearity,
\[
f(\pihat)
=
\frac1T
\sum_{t=0}^{T-1}
f(\pihat_t)
\]
and
\[
g(\pihat)
=
\frac1T
\sum_{t=0}^{T-1}
g(\pihat_t).
\]
Thus,
\begin{align}
f^*-f(\pihat)
+
\lambda
\left(
b'-g(\pihat)
\right)
\leq
\frac{
R^d(\lambda,T)
}{
T
},
\qquad
\lambda\in[0,U].
\label{eq:pd-master}
\end{align}

Define
\[
g_t
:=
g(\pihat_t)-b'.
\]
Since
\[
g(\pihat_t)\in[0,1],
\]
we have
\[
|g_t|
\leq
G_{b'}.
\]

Let
\[
\lambda_{t+1}'
:=
\mathbb P_{[0,U]}
\left[
\lambda_t-\eta g_t
\right].
\]
The rounding operator satisfies
\[
|\lambda_{t+1}-\lambda_{t+1}'|
\leq
\varepsilon_1.
\]
Therefore,
\begin{align*}
|\lambda_{t+1}-\lambda|^2
&\leq
\left(
\varepsilon_1+
|\lambda_{t+1}'-\lambda|
\right)^2
\\
&\leq
\varepsilon_1^2
+
2\varepsilon_1U
+
|\lambda_{t+1}'-\lambda|^2,
\end{align*}
where we used
\[
\lambda,\lambda_{t+1}'
\in
[0,U].
\]

Projection onto \([0,U]\) is non-expansive, so
\begin{align*}
|\lambda_{t+1}'-\lambda|^2
&\leq
|\lambda_t-\eta g_t-\lambda|^2
\\
&=
|\lambda_t-\lambda|^2
-
2\eta(\lambda_t-\lambda)g_t
+
\eta^2g_t^2
\\
&\leq
|\lambda_t-\lambda|^2
-
2\eta(\lambda_t-\lambda)g_t
+
\eta^2G_{b'}^2.
\end{align*}

Combining the preceding displays and rearranging gives
\[
(\lambda_t-\lambda)g_t
\leq
\frac{
\varepsilon_1^2+2\varepsilon_1U
}{
2\eta
}
+
\frac{
|\lambda_t-\lambda|^2
-
|\lambda_{t+1}-\lambda|^2
}{
2\eta
}
+
\frac{
\eta G_{b'}^2
}{
2
}.
\]

Summing from \(t=0\) to \(T-1\) yields
\begin{align*}
R^d(\lambda,T)
&\leq
T
\frac{
\varepsilon_1^2+2\varepsilon_1U
}{
2\eta
}
+
\frac{
|\lambda_0-\lambda|^2
-
|\lambda_T-\lambda|^2
}{
2\eta
}
+
\frac{
\eta G_{b'}^2T
}{
2
}
\\
&\leq
T
\frac{
\varepsilon_1^2+2\varepsilon_1U
}{
2\eta
}
+
\frac{
U^2
}{
2\eta
}
+
\frac{
\eta G_{b'}^2T
}{
2
}.
\end{align*}

Substituting
\[
\eta
=
\frac{U}{G_{b'}\sqrt T}
\]
gives
\begin{align}
\frac{
R^d(\lambda,T)
}{
T
}
\leq
G_{b'}\sqrt T
\frac{
\varepsilon_1^2+2\varepsilon_1U
}{
2U
}
+
\frac{
UG_{b'}
}{
\sqrt T
}
=:
\Gamma_T.
\label{eq:regretbound}
\end{align}

Taking \(\lambda=0\) in~\cref{eq:pd-master} gives
\[
f^*-f(\pihat)
\leq
\Gamma_T.
\]

Define
\[
v
:=
\left[
b'-g(\pihat)
\right]_+.
\]
If \(v=0\), the constraint conclusion follows immediately. Suppose
\(v>0\). Taking \(\lambda=U\) in~\cref{eq:pd-master} gives
\[
f^*-f(\pihat)+Uv
\leq
\Gamma_T.
\]

By strong duality,
\[
f^*
=
\max_{\pi\in\Pi_{\mathrm{mix}}}
\left\{
f(\pi)
+
\lambda^*
\left(
g(\pi)-b'
\right)
\right\}.
\]
Evaluating the maximum at \(\pihat\) yields
\[
f^*
\geq
f(\pihat)
+
\lambda^*
\left(
g(\pihat)-b'
\right)
=
f(\pihat)-\lambda^*v.
\]
Thus,
\[
f^*-f(\pihat)
\geq
-\lambda^*v.
\]
Combining the last two inequalities gives
\[
(U-\lambda^*)v
\leq
\Gamma_T.
\]

Since
\[
\varepsilon_1
\leq
U,
\]
we have
\[
\frac{
\varepsilon_1^2+2\varepsilon_1U
}{
2U
}
\leq
\frac{3\varepsilon_1}{2}.
\]
The assumed choice of \(\varepsilon_1\) gives
\[
G_{b'}\sqrt T
\frac{
\varepsilon_1^2+2\varepsilon_1U
}{
2U
}
\leq
\frac{
\varepsilon_{\mathrm{opt}}m
}{
2
}.
\]
The lower bound on \(T\) gives
\[
\frac{
UG_{b'}
}{
\sqrt T
}
\leq
\frac{
\varepsilon_{\mathrm{opt}}m
}{
2
}.
\]
Consequently,
\[
\Gamma_T
\leq
\varepsilon_{\mathrm{opt}}m.
\]

Since
\[
m
\leq
1,
\]
we obtain
\[
f^*-f(\pihat)
\leq
\varepsilon_{\mathrm{opt}}.
\]
Moreover,
\[
\frac{
\min\{1,U-\lambda^*\}
}{
U-\lambda^*
}
\leq
1,
\]
and hence
\[
\left[
b'-g(\pihat)
\right]_+
\leq
\varepsilon_{\mathrm{opt}}.
\]
\end{proof}

\begin{thmbox}
\begin{lemma}[Bounding the dual variable]
\label{lemma:dual-bound}
Suppose there exists
\(\bar\pi\in\Pi_{\mathrm{mix}}\) such that
\[
(1-\gamma)\widehat V_{c,\gamma}^{\bar\pi}(s)-b'
\geq
\sigma
\]
for some \(\sigma>0\). Then
\[
\lambda^*
\leq
\frac{
1+\omega
}{
\sigma
}.
\]
\end{lemma}
\end{thmbox}

\begin{proof}
By strong duality,
\[
(1-\gamma)\widehat V_{r_p,\gamma}^{\widehat\pi_\gamma^*}(s)
=
\max_{\pi\in\Pi_{\mathrm{mix}}}
\left\{
(1-\gamma)\widehat V_{r_p,\gamma}^{\pi}(s)
+
\lambda^*
\left(
(1-\gamma)\widehat V_{c,\gamma}^{\pi}(s)-b'
\right)
\right\}.
\]
Evaluating at \(\bar\pi\) gives
\[
(1-\gamma)\widehat V_{r_p,\gamma}^{\widehat\pi_\gamma^*}(s)
\geq
(1-\gamma)\widehat V_{r_p,\gamma}^{\bar\pi}(s)
+
\lambda^*\sigma.
\]
Since
\[
r_p
\in
[0,1+\omega],
\]
every normalized discounted \(r_p\)-value lies in
\([0,1+\omega]\). Therefore,
\[
\lambda^*\sigma
\leq
1+\omega.
\]
\end{proof}

\newpage
\section{Proof of~\cref{thm:ub-relaxed}}
\label{app:proof-relaxed}

\ubrelaxed*

\begin{proof}
The proof proceeds in four steps. Step 1 applies the
average-to-discounted bridge. Step 2 invokes the combined discounted
concentration event. Step 3 controls the empirical dual and
optimization errors. Step 4 applies the full suboptimality
decomposition.

Define
\[
\kappa_\gamma
:=
(1-\gamma)H.
\]
Under the relaxed parameter choice,
\[
\kappa_\gamma
\leq
\frac{\epsilon}{16}
=
\alpha_{\mathrm{rel}}.
\]

Let \(\mathcal E_{\mathrm{rel}}\) denote the event in
Theorem~\ref{thm:cross-discount-conc} with
\[
\alpha
=
\alpha_{\mathrm{rel}},
\]
covering:

\begin{enumerate}
\item every deterministic policy in the support of \(\piopt\);
\item the deterministic constraint-maximizing policy \(\pi_c^*\);
\item every empirical oracle policy
\(\pihat_\lambda\), \(\lambda\in\Lambda\).
\end{enumerate}

On this event,
\[
\left|
(1-\gamma)V_{q,\gamma}^{\pi}(s)
-
(1-\gamma)\widehat V_{q,\gamma}^{\pi}(s)
\right|
\leq
\alpha_{\mathrm{rel}}
\]
for \(q\in\{r,c\}\) and every relevant policy. By
external-mixture linearity, the same inequality holds for \(\piopt\)
and \(\pihat\).

Because
\[
\rho_c^{\pi_c^*}(s)
\geq
b,
\]
the bridge and concentration bounds imply
\begin{align*}
(1-\gamma)\widehat V_{c,\gamma}^{\pi_c^*}(s)
&\geq
(1-\gamma)V_{c,\gamma}^{\pi_c^*}(s)
-
\alpha_{\mathrm{rel}}
\\
&\geq
\rho_c^{\pi_c^*}(s)
-
\kappa_\gamma
-
\alpha_{\mathrm{rel}}
\\
&\geq
b-\frac{\epsilon}{8}.
\end{align*}
Since
\[
b'
=
b-\frac{\epsilon}{4},
\]
we have
\[
(1-\gamma)\widehat V_{c,\gamma}^{\pi_c^*}(s)-b'
\geq
\frac{\epsilon}{8}.
\]
Lemma~\ref{lemma:dual-bound} gives
\[
\lambda^*
\leq
\frac{
8(1+\omega)
}{
\epsilon
}
\leq
\frac9\epsilon.
\]
Since
\[
U
=
\frac{18}{\epsilon},
\]
we obtain
\[
U-\lambda^*
\geq
\frac9\epsilon
\geq
1.
\]

The average-optimal mixture \(\piopt\) is feasible for the empirical
discounted reference problem:
\begin{align*}
(1-\gamma)\widehat V_{c,\gamma}^{\piopt}(s)
&\geq
(1-\gamma)V_{c,\gamma}^{\piopt}(s)
-
\alpha_{\mathrm{rel}}
\\
&\geq
\rho_c^{\piopt}(s)
-
\kappa_\gamma
-
\alpha_{\mathrm{rel}}
\\
&\geq
b-\frac{\epsilon}{8}
\\
&\geq
b'.
\end{align*}

The parameter choices therefore satisfy the assumptions of
Theorem~\ref{thm:pd-guarantees}. Conditional on the empirical model,
\[
(1-\gamma)\widehat V_{r_p,\gamma}^{\pihat}(s)
\geq
(1-\gamma)\widehat V_{r_p,\gamma}^{\widehat\pi_\gamma^*}(s)
-
\frac{\epsilon}{4}
\]
and
\[
(1-\gamma)\widehat V_{c,\gamma}^{\pihat}(s)
\geq
b'
-
\frac{\epsilon}{4}.
\]

For the output constraint,
\begin{align*}
\rho_c^{\pihat}(s)
&\geq
(1-\gamma)V_{c,\gamma}^{\pihat}(s)-\kappa_\gamma
\\
&\geq
(1-\gamma)\widehat V_{c,\gamma}^{\pihat}(s)
-
\alpha_{\mathrm{rel}}
-
\kappa_\gamma
\\
&\geq
b'
-
\varepsilon_{\mathrm{opt}}
-
\alpha_{\mathrm{rel}}
-
\kappa_\gamma
\\
&=
b
-
\frac{\epsilon}{4}
-
\frac{\epsilon}{4}
-
\frac{\epsilon}{16}
-
\frac{\epsilon}{16}
\\
&=
b-\frac{5\epsilon}{8}
\\
&\geq
b-\epsilon.
\end{align*}

For reward optimality, use the complete decomposition
\begin{align*}
&
\rho_r^{\piopt}(s)-\rho_r^{\pihat}(s)
\\
={}&
\left(
\rho_r^{\piopt}(s)-(1-\gamma)V_{r,\gamma}^{\piopt}(s)
\right)
+
\left(
(1-\gamma)V_{r,\gamma}^{\piopt}(s)
-
(1-\gamma)\widehat V_{r,\gamma}^{\piopt}(s)
\right)
\\
&+
\left(
(1-\gamma)\widehat V_{r,\gamma}^{\piopt}(s)
-
(1-\gamma)\widehat V_{r_p,\gamma}^{\piopt}(s)
\right)
+
\left(
(1-\gamma)\widehat V_{r_p,\gamma}^{\piopt}(s)
-
(1-\gamma)\widehat V_{r_p,\gamma}^{\widehat\pi_\gamma^*}(s)
\right)
\\
&+
\left(
(1-\gamma)\widehat V_{r_p,\gamma}^{\widehat\pi_\gamma^*}(s)
-
(1-\gamma)\widehat V_{r_p,\gamma}^{\pihat}(s)
\right)
+
\left(
(1-\gamma)\widehat V_{r_p,\gamma}^{\pihat}(s)
-
(1-\gamma)\widehat V_{r,\gamma}^{\pihat}(s)
\right)
\\
&+
\left(
(1-\gamma)\widehat V_{r,\gamma}^{\pihat}(s)
-
(1-\gamma)V_{r,\gamma}^{\pihat}(s)
\right)
+
\left(
(1-\gamma)V_{r,\gamma}^{\pihat}(s)
-
\rho_r^{\pihat}(s)
\right).
\end{align*}

The first and last terms are at most \(\kappa_\gamma\), the second and
seventh terms are at most \(\alpha_{\mathrm{rel}}\), the third term is
nonpositive because \(r_p\geq r\), the fourth term is nonpositive by
feasibility and empirical optimality, the fifth term is at most
\(\varepsilon_{\mathrm{opt}}\), and the sixth term is at most
\(\omega\). Therefore,
\[
\rho_r^{\piopt}(s)-\rho_r^{\pihat}(s)
\leq
2\kappa_\gamma
+
2\alpha_{\mathrm{rel}}
+
\omega
+
\varepsilon_{\mathrm{opt}}
\leq
\frac{9\epsilon}{16}
<
\epsilon.
\]

It remains to verify the sample count. By
Theorem~\ref{thm:cross-discount-conc}, it suffices that
\[
n
\geq
C
\left(
\frac{B+H}{\alpha_{\mathrm{rel}}^2}
+
\frac1{\alpha_{\mathrm{rel}}(1-\gamma)}
\right)
\polylog(\cdots).
\]
Since
\[
\alpha_{\mathrm{rel}}
=
(B+H)(1-\gamma)
=
\frac{\epsilon}{16},
\]
both leading terms are of order
\[
\frac{B+H}{\epsilon^2}.
\]
Multiplying by \(SA\) completes the proof.
\end{proof}

\begin{thmbox}
\begin{lemma}[Relaxed suboptimality decomposition]
\label{lemma:decomposition-relaxed}
Suppose \(\piopt\) is feasible for the empirical discounted reference
CMDP. Suppose
\[
\left|
(1-\gamma)V_{r,\gamma}^{\piopt}(s)
-
(1-\gamma)\widehat V_{r,\gamma}^{\piopt}(s)
\right|
\leq
\alpha,
\]
\[
\left|
(1-\gamma)V_{r,\gamma}^{\pihat}(s)
-
(1-\gamma)\widehat V_{r,\gamma}^{\pihat}(s)
\right|
\leq
\alpha,
\]
and
\[
\left|
(1-\gamma)V_{c,\gamma}^{\pihat}(s)
-
(1-\gamma)\widehat V_{c,\gamma}^{\pihat}(s)
\right|
\leq
\alpha.
\]
Suppose also that
\[
\left|
(1-\gamma)V_{q,\gamma}^{\pi}(s)
-
\rho_q^\pi(s)
\right|
\leq
\kappa
\]
for \(q\in\{r,c\}\) and every relevant policy, and that
\[
(1-\gamma)\widehat V_{r_p,\gamma}^{\pihat}(s)
\geq
(1-\gamma)\widehat V_{r_p,\gamma}^{\widehat\pi_\gamma^*}(s)
-
\varepsilon_{\mathrm{opt}},
\]
\[
(1-\gamma)\widehat V_{c,\gamma}^{\pihat}(s)
\geq
b'
-
\varepsilon_{\mathrm{opt}}.
\]
Then
\[
\rho_r^{\piopt}(s)
-
\rho_r^{\pihat}(s)
\leq
2\kappa
+
2\alpha
+
\omega
+
\varepsilon_{\mathrm{opt}},
\]
and
\[
\rho_c^{\pihat}(s)
\geq
b'
-
\varepsilon_{\mathrm{opt}}
-
\alpha
-
\kappa.
\]
\end{lemma}
\end{thmbox}

\begin{proof}
For the constraint,
\begin{align*}
\rho_c^{\pihat}(s)
&\geq
(1-\gamma)V_{c,\gamma}^{\pihat}(s)-\kappa
\\
&\geq
(1-\gamma)\widehat V_{c,\gamma}^{\pihat}(s)-\alpha-\kappa
\\
&\geq
b'
-
\varepsilon_{\mathrm{opt}}
-
\alpha
-
\kappa.
\end{align*}

For reward,
\begin{align*}
\rho_r^{\piopt}(s)
&\leq
(1-\gamma)V_{r,\gamma}^{\piopt}(s)+\kappa
\\
&\leq
(1-\gamma)\widehat V_{r,\gamma}^{\piopt}(s)+\alpha+\kappa
\\
&\leq
(1-\gamma)\widehat V_{r_p,\gamma}^{\piopt}(s)+\alpha+\kappa
\\
&\leq
(1-\gamma)\widehat V_{r_p,\gamma}^{\widehat\pi_\gamma^*}(s)
+
\alpha+\kappa
\\
&\leq
(1-\gamma)\widehat V_{r_p,\gamma}^{\pihat}(s)
+
\varepsilon_{\mathrm{opt}}
+
\alpha+\kappa
\\
&\leq
(1-\gamma)\widehat V_{r,\gamma}^{\pihat}(s)
+
\omega
+
\varepsilon_{\mathrm{opt}}
+
\alpha+\kappa
\\
&\leq
(1-\gamma)V_{r,\gamma}^{\pihat}(s)
+
2\alpha
+
\omega
+
\varepsilon_{\mathrm{opt}}
+
\kappa
\\
&\leq
\rho_r^{\pihat}(s)
+
2\kappa
+
2\alpha
+
\omega
+
\varepsilon_{\mathrm{opt}}.
\end{align*}
\end{proof}

\newpage
\section{Proof of~\cref{thm:ub-strict}}
\label{app:proof-strict}

\ubstrict*

\begin{proof}
Define
\[
\kappa_\gamma
:=
(1-\gamma)H.
\]
Under the strict parameter choice,
\[
\kappa_\gamma
\leq
\frac{\epsilon\zeta}{128}
=
\alpha_{\mathrm{str}}.
\]

Let \(\mathcal E_{\mathrm{str}}\) denote the event in
Theorem~\ref{thm:cross-discount-conc} with
\[
\alpha
=
\alpha_{\mathrm{str}},
\]
covering the deterministic policies in the support of \(\piopt\),
\(\pi_c^*\), and every empirical oracle policy.

For the constraint-maximizing policy,
\[
\rho_c^{\pi_c^*}(s)
\geq
b+\zeta.
\]
Therefore,
\begin{align*}
(1-\gamma)\widehat V_{c,\gamma}^{\pi_c^*}(s)
&\geq
(1-\gamma)V_{c,\gamma}^{\pi_c^*}(s)
-
\alpha_{\mathrm{str}}
\\
&\geq
\rho_c^{\pi_c^*}(s)
-
\kappa_\gamma
-
\alpha_{\mathrm{str}}
\\
&\geq
b+\zeta-\kappa_\gamma-\alpha_{\mathrm{str}}.
\end{align*}
Hence,
\begin{align*}
(1-\gamma)\widehat V_{c,\gamma}^{\pi_c^*}(s)-b'
&\geq
\zeta-\Delta-\kappa_\gamma-\alpha_{\mathrm{str}}
\\
&\geq
\frac{\zeta}{2}.
\end{align*}
Lemma~\ref{lemma:dual-bound} gives
\[
\lambda^*
\leq
\frac{2(1+\omega)}{\zeta}
\leq
\frac3\zeta.
\]
Since
\[
U
=
\frac8\zeta,
\]
we have
\[
U-\lambda^*
\geq
\frac5\zeta
\geq
1.
\]

The primal--dual theorem gives
\[
(1-\gamma)\widehat V_{r_p,\gamma}^{\pihat}(s)
\geq
(1-\gamma)\widehat V_{r_p,\gamma}^{\widehat\pi_\gamma^*}(s)
-
\varepsilon_{\mathrm{opt}}
\]
and
\[
(1-\gamma)\widehat V_{c,\gamma}^{\pihat}(s)
\geq
b+\Delta-\varepsilon_{\mathrm{opt}}.
\]

For feasibility,
\begin{align*}
\rho_c^{\pihat}(s)
&\geq
(1-\gamma)V_{c,\gamma}^{\pihat}(s)-\kappa_\gamma
\\
&\geq
(1-\gamma)\widehat V_{c,\gamma}^{\pihat}(s)
-
\alpha_{\mathrm{str}}
-
\kappa_\gamma
\\
&\geq
b+\Delta
-
\varepsilon_{\mathrm{opt}}
-
\alpha_{\mathrm{str}}
-
\kappa_\gamma
\\
&\geq
b.
\end{align*}

Define
\[
b_0
:=
b-\kappa_\gamma-\alpha_{\mathrm{str}},
\]
and let
\[
\widetilde\pi_\gamma^*
\in
\argmax_{\pi\in\Pi_{\mathrm{mix}}}
(1-\gamma)\widehat V_{r_p,\gamma}^{\pi}(s)
\quad
\text{s.t.}
\quad
(1-\gamma)\widehat V_{c,\gamma}^{\pi}(s)
\geq
b_0.
\]
The true average-optimal policy is feasible for this auxiliary problem:
\[
(1-\gamma)\widehat V_{c,\gamma}^{\piopt}(s)
\geq
\rho_c^{\piopt}(s)
-
\kappa_\gamma
-
\alpha_{\mathrm{str}}
\geq
b_0.
\]

The threshold difference is
\[
d_b
:=
b'-b_0
=
\Delta+\kappa_\gamma+\alpha_{\mathrm{str}}
\leq
\frac{
6\epsilon\zeta
}{
128
}.
\]
By Lemma~\ref{lemma:sensitivity},
\[
(1-\gamma)\widehat V_{r_p,\gamma}^{\widetilde\pi_\gamma^*}(s)
-
(1-\gamma)\widehat V_{r_p,\gamma}^{\widehat\pi_\gamma^*}(s)
\leq
d_b\lambda^*
\leq
\frac{18\epsilon}{128}.
\]

Use the full strict suboptimality decomposition:
\begin{align*}
&
\rho_r^{\piopt}(s)-\rho_r^{\pihat}(s)
\\
={}&
\left(
\rho_r^{\piopt}(s)-(1-\gamma)V_{r,\gamma}^{\piopt}(s)
\right)
+
\left(
(1-\gamma)V_{r,\gamma}^{\piopt}(s)
-
(1-\gamma)\widehat V_{r,\gamma}^{\piopt}(s)
\right)
\\
&+
\left(
(1-\gamma)\widehat V_{r,\gamma}^{\piopt}(s)
-
(1-\gamma)\widehat V_{r_p,\gamma}^{\piopt}(s)
\right)
+
\left(
(1-\gamma)\widehat V_{r_p,\gamma}^{\piopt}(s)
-
(1-\gamma)\widehat V_{r_p,\gamma}^{\widetilde\pi_\gamma^*}(s)
\right)
\\
&+
\left(
(1-\gamma)\widehat V_{r_p,\gamma}^{\widetilde\pi_\gamma^*}(s)
-
(1-\gamma)\widehat V_{r_p,\gamma}^{\widehat\pi_\gamma^*}(s)
\right)
\\
&+
\left(
(1-\gamma)\widehat V_{r_p,\gamma}^{\widehat\pi_\gamma^*}(s)
-
(1-\gamma)\widehat V_{r_p,\gamma}^{\pihat}(s)
\right)
\\
&+
\left(
(1-\gamma)\widehat V_{r_p,\gamma}^{\pihat}(s)
-
(1-\gamma)\widehat V_{r,\gamma}^{\pihat}(s)
\right)
+
\left(
(1-\gamma)\widehat V_{r,\gamma}^{\pihat}(s)
-
(1-\gamma)V_{r,\gamma}^{\pihat}(s)
\right)
\\
&+
\left(
(1-\gamma)V_{r,\gamma}^{\pihat}(s)
-
\rho_r^{\pihat}(s)
\right).
\end{align*}

The first and last terms are bounded by \(\kappa_\gamma\), the second
and eighth by \(\alpha_{\mathrm{str}}\), the third and fourth are
nonpositive, the fifth is the sensitivity error, the sixth is the
primal--dual optimization error, and the seventh is at most \(\omega\).
Thus,
\begin{align*}
\rho_r^{\piopt}(s)-\rho_r^{\pihat}(s)
&\leq
2\kappa_\gamma
+
2\alpha_{\mathrm{str}}
+
\omega
+
\varepsilon_{\mathrm{opt}}
+
d_b\lambda^*
\\
&\leq
\frac{2\epsilon\zeta}{128}
+
\frac{2\epsilon\zeta}{128}
+
\frac{\epsilon}{16}
+
\frac{\epsilon\zeta}{128}
+
\frac{18\epsilon}{128}
\\
&<
\epsilon.
\end{align*}

Finally,
\[
n
=
\widetilde O
\left(
\frac{
B+H
}{
\epsilon^2\zeta^2
}
\right),
\]
and multiplying by \(SA\) completes the proof.
\end{proof}

\begin{thmbox}
\begin{lemma}[Strict suboptimality decomposition]
\label{lemma:decomposition-strict}
Let \(b_0<b'\), and let
\(\widetilde\pi_\gamma^*\) and \(\widehat\pi_\gamma^*\) be optimal
policies for the empirical normalized discounted CMDPs with thresholds
\(b_0\) and \(b'\), respectively. Suppose \(\piopt\) is feasible for
the problem with threshold \(b_0\).

Suppose
\[
\left|
(1-\gamma)V_{r,\gamma}^{\piopt}(s)
-
(1-\gamma)\widehat V_{r,\gamma}^{\piopt}(s)
\right|
\leq
\alpha,
\]
\[
\left|
(1-\gamma)V_{r,\gamma}^{\pihat}(s)
-
(1-\gamma)\widehat V_{r,\gamma}^{\pihat}(s)
\right|
\leq
\alpha,
\]
and suppose the average-to-discounted bridge error is at most
\(\kappa\) for \(\piopt\) and \(\pihat\). Suppose further that
\[
(1-\gamma)\widehat V_{r_p,\gamma}^{\pihat}(s)
\geq
(1-\gamma)\widehat V_{r_p,\gamma}^{\widehat\pi_\gamma^*}(s)
-
\varepsilon_{\mathrm{opt}}.
\]

If \(\lambda^*\) is an optimal dual multiplier for the problem with
threshold \(b'\), then
\[
\rho_r^{\piopt}(s)
-
\rho_r^{\pihat}(s)
\leq
2\kappa
+
2\alpha
+
\omega
+
\varepsilon_{\mathrm{opt}}
+
(b'-b_0)\lambda^*.
\]
\end{lemma}
\end{thmbox}

\begin{proof}
Since \(\piopt\) is feasible for the problem with threshold \(b_0\),
\[
(1-\gamma)\widehat V_{r_p,\gamma}^{\widetilde\pi_\gamma^*}(s)
\geq
(1-\gamma)\widehat V_{r_p,\gamma}^{\piopt}(s).
\]
By Lemma~\ref{lemma:sensitivity},
\[
(1-\gamma)\widehat V_{r_p,\gamma}^{\widetilde\pi_\gamma^*}(s)
-
(1-\gamma)\widehat V_{r_p,\gamma}^{\widehat\pi_\gamma^*}(s)
\leq
(b'-b_0)\lambda^*.
\]
Therefore,
\begin{align*}
\rho_r^{\piopt}(s)
&\leq
(1-\gamma)V_{r,\gamma}^{\piopt}(s)+\kappa
\\
&\leq
(1-\gamma)\widehat V_{r,\gamma}^{\piopt}(s)+\alpha+\kappa
\\
&\leq
(1-\gamma)\widehat V_{r_p,\gamma}^{\piopt}(s)+\alpha+\kappa
\\
&\leq
(1-\gamma)\widehat V_{r_p,\gamma}^{\widetilde\pi_\gamma^*}(s)
+\alpha+\kappa
\\
&\leq
(1-\gamma)\widehat V_{r_p,\gamma}^{\widehat\pi_\gamma^*}(s)
+
(b'-b_0)\lambda^*
+
\alpha+\kappa
\\
&\leq
(1-\gamma)\widehat V_{r_p,\gamma}^{\pihat}(s)
+
\varepsilon_{\mathrm{opt}}
+
(b'-b_0)\lambda^*
+
\alpha+\kappa
\\
&\leq
(1-\gamma)\widehat V_{r,\gamma}^{\pihat}(s)
+
\omega
+
\varepsilon_{\mathrm{opt}}
+
(b'-b_0)\lambda^*
+
\alpha+\kappa
\\
&\leq
(1-\gamma)V_{r,\gamma}^{\pihat}(s)
+
2\alpha
+
\omega
+
\varepsilon_{\mathrm{opt}}
+
(b'-b_0)\lambda^*
+
\kappa
\\
&\leq
\rho_r^{\pihat}(s)
+
2\kappa
+
2\alpha
+
\omega
+
\varepsilon_{\mathrm{opt}}
+
(b'-b_0)\lambda^*.
\end{align*}
\end{proof}

\newpage
\section{Discounted Concentration Inequalities}
\label{app:concentration}

This appendix proves Theorem~\ref{thm:cross-discount-conc}. The proof has four parts. First, a deterministic bridge compares average and normalized discounted values. Second, we record the instance-dependent fixed-policy plug-in inequality. Third, reward perturbation and the state--action absorbing construction reduce each data-dependent oracle policy to row-independent candidate families. Fourth, conditional Bernstein inequalities verify the hierarchy needed to retain the propagated-variance term in the plug-in analysis.

\begin{thmbox}
\begin{lemma}[Policy-wise average-to-discounted bridge]
\label{lemma:average-discount-bridge}
For every $\pi\in\Pi_{\mathrm{det}}$, $q\in\{r,c\}$, and $\gamma\in(0,1)$,
\begin{align}
\left\|
(1-\gamma)V_{q,\gamma}^{\pi}-\rho_q^\pi
\right\|_\infty
\leq
(1-\gamma)\spannorm{h_q^\pi}
\leq
(1-\gamma)H.
\label{eq:average-discount-bridge}
\end{align}
The same conclusion holds for every external mixture of policies from $\Pi_{\mathrm{det}}$.
\end{lemma}
\end{thmbox}

\begin{proof}
The Poisson equation gives $q_\pi=\rho_q^\pi+(I-P_\pi)h_q^\pi$. Since $V_{q,\gamma}^\pi=(I-\gamma P_\pi)^{-1}q_\pi$ and $(I-\gamma P_\pi)^{-1}\rho_q^\pi=(1-\gamma)^{-1}\rho_q^\pi$,
\[
(1-\gamma)V_{q,\gamma}^{\pi}-\rho_q^\pi
=
(1-\gamma)(I-\gamma P_\pi)^{-1}(I-P_\pi)h_q^\pi.
\]
The Neumann series yields
\[
(I-\gamma P_\pi)^{-1}(I-P_\pi)h_q^\pi
=
h_q^\pi-(1-\gamma)\sum_{t=1}^{\infty}\gamma^{t-1}P_\pi^t h_q^\pi.
\]
After shifting $h_q^\pi$ by a constant, its entries lie in $[0,\spannorm{h_q^\pi}]$. The second term is a convex combination of vectors in the same interval, proving~\cref{eq:average-discount-bridge}. External-mixture linearity proves the last statement.
\end{proof}

\begin{thmbox}
\begin{lemma}[Fixed-policy normalized discounted concentration]
\label{lemma:fixed-discounted-conc}
Let $\pi\in\Pi_{\mathrm{det}}$ be independent of the transition samples, let $q\in\{r,c\}$, and suppose Assumption~\ref{assum:policywise-variance} holds with $B+H\leq(1-\gamma)^{-1}$. There is a universal $C>0$ such that, if
\[
n\geq C\left(\frac{B+H}{\alpha^2}+\frac1{\alpha(1-\gamma)}\right)
\polylog\left(S,A,\frac1\delta,\frac1\alpha,\frac1{1-\gamma},B+H\right),
\]
then, with probability at least $1-\delta$,
\[
(1-\gamma)
\left\|V_{q,\gamma}^{\pi}-\widehat V_{q,\gamma}^{\pi}\right\|_\infty
\leq\alpha.
\]
\end{lemma}
\end{thmbox}

\begin{proof}
The instance-dependent fixed-policy plug-in inequality of \citet[Lemma~1]{li2020breaking}, with logarithmic factors denoted by $\ell$, gives
\begin{align}
\left\|V_{q,\gamma}^{\pi}-\widehat V_{q,\gamma}^{\pi}\right\|_\infty
\leq
C\left[
\sqrt{\frac\ell n}\,
\gamma\left\|(I-\gamma P_\pi)^{-1}\sigma_{q,\gamma}^{\pi}\right\|_\infty
+
\frac{\ell}{n(1-\gamma)^2}
\right].
\label{eq:fixed-plugin}
\end{align}
Multiplication by $1-\gamma$ and Assumption~\ref{assum:policywise-variance} imply
\[
(1-\gamma)
\left\|V_{q,\gamma}^{\pi}-\widehat V_{q,\gamma}^{\pi}\right\|_\infty
\leq
C\left[
\sqrt{\frac{(B+H)\ell}{n}}
+
\frac{\ell}{n(1-\gamma)}
\right].
\]
The displayed sample size makes both terms at most a constant fraction of $\alpha$.
\end{proof}

\begin{thmbox}
\begin{lemma}[Simultaneous perturbed action gap]
\label{lemma:discounted-gap}
There is a universal $c_{\mathrm{gap}}>0$ such that, with probability at least $1-\delta/4$, simultaneously for every $\lambda\in\Lambda$, the empirical discounted MDP with reward $r_p+\lambda c$ has a unique optimal action in every state and action gap at least
\begin{align}
\iota
=
c_{\mathrm{gap}}
\frac{\omega\delta(1-\gamma)}{|\Lambda|SA^2}.
\label{eq:gap-size}
\end{align}
\end{lemma}
\end{thmbox}

\begin{proof}
Apply the simultaneous random-perturbation gap result of \citet[Lemma~16]{vaswani_near-optimal_2022} to the finite grid $\Lambda$, with the failure probability divided by a universal constant. Scaling the Lagrangian reward by $1+\lambda$ does not change its optimal actions; its range affects only logarithmic factors later in the proof.
\end{proof}

For a row $z=(x,a)$, write $\mathcal D_z$ for the $n$ transition samples used to form $\widehat P(\cdot\mid z)$ and $\mathcal D_{-z}$ for all remaining samples.

Set
\[
\overline R:=1+U+\omega
\]
and let $\iota$ be the deterministic gap parameter in
Lemma~\ref{lemma:discounted-gap}. Define the deterministic cardinality
bound
\[
\overline K
:=
1+
\left\lceil
\frac{2(1+\gamma)\overline R}
{\iota(1-\gamma)^2}
\right\rceil.
\]

\begin{thmbox}
\begin{lemma}[Row-independent oracle candidates]
\label{lemma:row-candidates}
For every $z$ and $\lambda\in\Lambda$, there exists a finite
collection $\Pi_{z,\lambda}\subset\Pi_{\mathrm{det}}$ such that:
\begin{enumerate}
\item $\Pi_{z,\lambda}$ is measurable with respect to
$(\mathcal D_{-z},Z)$ and is therefore independent of $\mathcal D_z$;
\item on the gap event in Lemma~\ref{lemma:discounted-gap}, the
empirical oracle policy satisfies
\[
\pihat_\lambda\in\Pi_{z,\lambda};
\]
\item uniformly over $z$ and $\lambda$,
\[
|\Pi_{z,\lambda}|
\leq
\overline K.
\]
Consequently, for a universal constant $C>0$,
\[
\log|\Pi_{z,\lambda}|
\leq
C\log\left(
\frac{SA|\Lambda|(1+U+\omega)}
{\delta\omega(1-\gamma)^3}
\right).
\]
\end{enumerate}
\end{lemma}
\end{thmbox}

\begin{proof}
Fix $z=(x,a)$ and $\lambda\in\Lambda$, and write
\[
\alpha_\lambda:=r_p+\lambda c.
\]
By the boundedness assumptions on $r_p$, $c$, and $\lambda$,
\[
0\leq \alpha_\lambda(s,a')\leq\overline R
\]
for every state--action pair $(s,a')$.

Following the state--action absorbing construction of
\citet[Lemma~5]{vaswani_near-optimal_2022}, replace the empirical
transition row $z=(x,a)$ by the deterministic self-loop at $x$, and
replace the one-step reward at $z$ by a scalar $u$. All other
transition rows and rewards are left unchanged. Denote the resulting
auxiliary empirical discounted MDP by
$\widehat M_{z,\lambda,u}$.

Let
\[
\widehat V_{z,\lambda,u}^*,
\qquad
\widehat Q_{z,\lambda,u}^*
\]
be its optimal value and state--action value functions, and let
$\pi_{z,\lambda,u}$ be its lexicographically selected optimal
deterministic policy.

For the original empirical MDP, define
\[
u_z^*
:=
\widehat Q_{\alpha_\lambda,\gamma}^*(x,a)
-
\gamma
\widehat V_{\alpha_\lambda,\gamma}^*(x).
\]
Since
\[
0
\leq
\widehat V_{\alpha_\lambda,\gamma}^*
\leq
\frac{\overline R}{1-\gamma}\mathbf 1,
\qquad
0
\leq
\widehat Q_{\alpha_\lambda,\gamma}^*
\leq
\frac{\overline R}{1-\gamma}\mathbf 1,
\]
we have
\[
-\frac{\gamma\overline R}{1-\gamma}
\leq
u_z^*
\leq
\frac{\overline R}{1-\gamma}.
\]
Thus $u_z^*$ belongs to the deterministic interval
\[
\mathcal I
:=
\left[
-\frac{\gamma\overline R}{1-\gamma},
\frac{\overline R}{1-\gamma}
\right].
\]

At the exact value $u=u_z^*$, the optimal Bellman equations of the
auxiliary MDP coincide with those of the original empirical MDP. In
particular, at the replaced state--action pair,
\[
u_z^*
+
\gamma
\widehat V_{\alpha_\lambda,\gamma}^*(x)
=
\widehat Q_{\alpha_\lambda,\gamma}^*(x,a),
\]
while all other Bellman equations are unchanged. Hence
\[
\widehat V_{z,\lambda,u_z^*}^*
=
\widehat V_{\alpha_\lambda,\gamma}^*,
\qquad
\widehat Q_{z,\lambda,u_z^*}^*
=
\widehat Q_{\alpha_\lambda,\gamma}^*.
\]
On the gap event, the optimal action at every state is unique, and
therefore
\[
\pi_{z,\lambda,u_z^*}
=
\pihat_\lambda.
\]

Construct a deterministic grid $\mathcal G\subset\mathcal I$ that
contains both endpoints of $\mathcal I$ and has maximum spacing at
most
\[
h
:=
\frac{\iota(1-\gamma)}{2}.
\]
Define
\[
\Pi_{z,\lambda}
:=
\left\{
\pi_{z,\lambda,u}:u\in\mathcal G
\right\}.
\]

Every auxiliary MDP used to construct $\Pi_{z,\lambda}$ replaces row
$z$ by a fixed deterministic self-loop. Hence its optimizer depends
only on the remaining empirical rows and on the reward perturbation.
It follows that $\Pi_{z,\lambda}$ is measurable with respect to
$(\mathcal D_{-z},Z)$ and is independent of $\mathcal D_z$.

Let $\bar u_z\in\mathcal G$ be a closest grid point to $u_z^*$. Since
the maximum grid spacing is at most $h$,
\[
|\bar u_z-u_z^*|
\leq
\frac{h}{2}
=
\frac{\iota(1-\gamma)}{4}.
\]
Discounted optimal-value stability therefore gives
\[
\left\|
\widehat Q_{z,\lambda,\bar u_z}^*
-
\widehat Q_{z,\lambda,u_z^*}^*
\right\|_\infty
\leq
\frac{|\bar u_z-u_z^*|}{1-\gamma}
\leq
\frac{\iota}{4}.
\]

On the gap event, for every state $s$ and every
$a'\neq\pihat_\lambda(s)$,
\[
\widehat Q_{\alpha_\lambda,\gamma}^*
\bigl(s,\pihat_\lambda(s)\bigr)
-
\widehat Q_{\alpha_\lambda,\gamma}^*(s,a')
\geq
\iota.
\]
Consequently,
\begin{align*}
&
\widehat Q_{z,\lambda,\bar u_z}^*
\bigl(s,\pihat_\lambda(s)\bigr)
-
\widehat Q_{z,\lambda,\bar u_z}^*(s,a')
\\
&\qquad\geq
\iota
-
2\left\|
\widehat Q_{z,\lambda,\bar u_z}^*
-
\widehat Q_{z,\lambda,u_z^*}^*
\right\|_\infty
\geq
\frac{\iota}{2}.
\end{align*}
Thus the optimal action at every state remains unchanged, and
\[
\pi_{z,\lambda,\bar u_z}
=
\pi_{z,\lambda,u_z^*}
=
\pihat_\lambda.
\]
Since $\bar u_z\in\mathcal G$, this proves
\[
\pihat_\lambda\in\Pi_{z,\lambda}
\]
on the gap event.

Finally, the length of $\mathcal I$ is
\[
|\mathcal I|
=
\frac{(1+\gamma)\overline R}{1-\gamma}.
\]
Therefore,
\[
|\Pi_{z,\lambda}|
\leq
|\mathcal G|
\leq
1+
\left\lceil
\frac{|\mathcal I|}{h}
\right\rceil
=
1+
\left\lceil
\frac{2(1+\gamma)\overline R}
{\iota(1-\gamma)^2}
\right\rceil
=
\overline K.
\]
Substituting the value of $\iota$ from
Lemma~\ref{lemma:discounted-gap} and absorbing polynomial powers of
$S$ and $A$ inside the logarithm into a universal constant gives
\[
\log|\Pi_{z,\lambda}|
\leq
C\log\left(
\frac{SA|\Lambda|(1+U+\omega)}
{\delta\omega(1-\gamma)^3}
\right).
\]
\end{proof}

For a deterministic policy $\pi$ and an evaluation signal $q$, define
the true Bernstein hierarchy by
\begin{align*}
V_{q,\gamma}^{\pi,(0)}
&:=
V_{q,\gamma}^{\pi},\\
r_{q,\gamma}^{\pi,(j)}
&:=
\sqrt{
\operatorname{Var}_{P_\pi}
\bigl(V_{q,\gamma}^{\pi,(j-1)}\bigr)
},
\qquad j\geq1,\\
V_{q,\gamma}^{\pi,(j)}
&:=
(I-\gamma P_\pi)^{-1}
r_{q,\gamma}^{\pi,(j)},
\qquad j\geq1.
\end{align*}
Here and below, variances and square roots are taken componentwise.
Let
\[
L
:=
\left\lceil
\log\frac{e}{1-\gamma}
\right\rceil,
\]
where $\log$ denotes the natural logarithm.

\begin{thmbox}
\begin{lemma}[Conditional Bernstein hierarchy]
\label{lemma:bernstein-hierarchy}
Define the deterministic logarithmic factor
\[
\ell
:=
\max\left\{
1,\,
\log\left(
\frac{
8SA|\Lambda|(L+1)\overline K
}{
\delta
}
\right)
\right\}.
\]
There is an event $\mathcal E_{\mathrm{hier}}$ of probability at least
$1-\delta/2$ on which, simultaneously for every row $z$,
$\lambda\in\Lambda$, $q\in\{r,c\}$,
$\pi\in\Pi_{z,\lambda}$, and $0\leq j\leq L$,
\begin{align}
\left|
(\widehat P_z-P_z)
V_{q,\gamma}^{\pi,(j)}
\right|
\leq
\sqrt{
\frac{2\ell}{n}
\operatorname{Var}_{P_z}
\bigl(V_{q,\gamma}^{\pi,(j)}\bigr)
}
+
\frac{2\ell}{3n}
\spannorm{V_{q,\gamma}^{\pi,(j)}}.
\label{eq:hierarchy-row-bound}
\end{align}
Moreover, $\ell$ is logarithmic in the parameters appearing in
Theorem~\ref{thm:cross-discount-conc}.
\end{lemma}
\end{thmbox}

\begin{proof}
Fix a row $z$ and condition on $(\mathcal D_{-z},Z)$. Under this
conditioning, every policy $\pi\in\Pi_{z,\lambda}$ is fixed and is
independent of $\mathcal D_z$. Since the true transition kernel is
nonrandom, every corresponding hierarchy vector
$V_{q,\gamma}^{\pi,(j)}$ is also fixed under this conditioning.

For a fixed tuple $(z,\lambda,q,\pi,j)$, let
\[
X_1,\ldots,X_n
\stackrel{\mathrm{i.i.d.}}{\sim}
P_z
\]
be the next-state samples in $\mathcal D_z$. Then
\[
(\widehat P_z-P_z)V_{q,\gamma}^{\pi,(j)}
=
\frac1n\sum_{i=1}^n
\left(
V_{q,\gamma}^{\pi,(j)}(X_i)
-
P_zV_{q,\gamma}^{\pi,(j)}
\right).
\]
The summands have variance
\[
\operatorname{Var}_{P_z}
\bigl(V_{q,\gamma}^{\pi,(j)}\bigr)
\]
and absolute value bounded by
\[
\spannorm{V_{q,\gamma}^{\pi,(j)}}.
\]
The two-sided Bernstein inequality therefore gives
\cref{eq:hierarchy-row-bound} with conditional failure probability at
most
\[
2e^{-\ell}.
\]

For a fixed row $z$, take a union bound over
$\lambda\in\Lambda$, the two evaluation signals, the $L+1$ hierarchy
levels, and all policies in $\Pi_{z,\lambda}$. Since
\[
|\Pi_{z,\lambda}|
\leq
\overline K
\]
deterministically, integrating over $(\mathcal D_{-z},Z)$ gives the
same unconditional failure bound.

A final union bound over all $SA$ rows gives total failure probability
at most
\[
2e^{-\ell}
\cdot
2SA|\Lambda|(L+1)\overline K
\leq
\frac{\delta}{2}.
\]
This proves the simultaneous statement.

Finally, Lemma~\ref{lemma:row-candidates} implies
\[
\log\overline K
\leq
C\log\left(
\frac{SA|\Lambda|(1+U+\omega)}
{\delta\omega(1-\gamma)^3}
\right),
\]
so $\ell$ has the claimed logarithmic dependence.
\end{proof}

\begin{thmbox}
\begin{lemma}[Hierarchy-to-plug-in bound]
\label{lemma:hierarchy-plugin}
Let $\pi$ be a possibly data-dependent deterministic policy, and let
$q\in[0,1]$. Suppose that, for every state $x$ and every
$0\leq j\leq L$, \cref{eq:hierarchy-row-bound} holds with
\[
z_x:=(x,\pi(x)).
\]
If
\[
n
\geq
\frac{C\ell}{1-\gamma},
\]
then
\begin{align}
\left\|
V_{q,\gamma}^{\pi}
-
\widehat V_{q,\gamma}^{\pi}
\right\|_\infty
\leq
C\left[
\sqrt{\frac{\ell}{n}}\,
\gamma
\left\|
(I-\gamma P_\pi)^{-1}
\sigma_{q,\gamma}^{\pi}
\right\|_\infty
+
\frac{\ell}{n(1-\gamma)^2}
\right],
\label{eq:data-dependent-plugin}
\end{align}
where
\[
\sigma_{q,\gamma}^{\pi}
:=
\sqrt{
\operatorname{Var}_{P_\pi}
\bigl(V_{q,\gamma}^{\pi}\bigr)
}.
\]
\end{lemma}
\end{thmbox}

\begin{proof}
For every state $x$, the $x$-th row of
$\widehat P_\pi-P_\pi$ is
\[
\widehat P_{z_x}-P_{z_x},
\qquad
z_x=(x,\pi(x)).
\]
Consequently, stacking the assumed rowwise inequalities over all
states gives, componentwise, for every $0\leq j\leq L$,
\begin{align}
\left|
(\widehat P_\pi-P_\pi)
V_{q,\gamma}^{\pi,(j)}
\right|
\leq&
\sqrt{\frac{2\ell}{n}}\,
\sqrt{
\operatorname{Var}_{P_\pi}
\bigl(V_{q,\gamma}^{\pi,(j)}\bigr)
}
\nonumber\\
&+
\frac{2\ell}{3n}
\spannorm{V_{q,\gamma}^{\pi,(j)}}
\mathbf 1.
\label{eq:componentwise-hierarchy}
\end{align}

Because $q\in[0,1]$,
\[
V_{q,\gamma}^{\pi,(0)}
=
(I-\gamma P_\pi)^{-1}q_\pi
\geq0.
\]
Moreover,
\[
r_{q,\gamma}^{\pi,(j)}\geq0
\]
and $(I-\gamma P_\pi)^{-1}$ is entrywise nonnegative. Hence
\[
V_{q,\gamma}^{\pi,(j)}\geq0
\]
for every $j\geq1$. It follows that
\[
\spannorm{V_{q,\gamma}^{\pi,(j)}}
\leq
\left\|
V_{q,\gamma}^{\pi,(j)}
\right\|_\infty.
\]

Set
\[
\beta_1:=2\ell.
\]
Since
\[
\frac{2\ell}{3n}
\leq
\frac{\beta_1}{n},
\]
\cref{eq:componentwise-hierarchy} implies, componentwise,
\begin{align}
\left|
(\widehat P_\pi-P_\pi)
V_{q,\gamma}^{\pi,(j)}
\right|
\leq&
\sqrt{\frac{\beta_1}{n}}\,
\sqrt{
\operatorname{Var}_{P_\pi}
\bigl(V_{q,\gamma}^{\pi,(j)}\bigr)
}
\nonumber\\
&+
\frac{\beta_1}{n}
\left\|
V_{q,\gamma}^{\pi,(j)}
\right\|_\infty
\mathbf 1,
\qquad
0\leq j\leq L.
\label{eq:li-hierarchy-condition}
\end{align}
This is precisely the Bernstein hierarchy condition used in
\citet[Lemma~2, Eq.~(75)]{li2020breaking}.

For completeness, define the common one-step signals
\[
b_{q,\gamma}^{\pi,(0)}
:=
q_\pi,
\qquad
b_{q,\gamma}^{\pi,(j)}
:=
r_{q,\gamma}^{\pi,(j)},
\quad j\geq1,
\]
and define
\[
\widehat V_{q,\gamma}^{\pi,(j)}
:=
(I-\gamma\widehat P_\pi)^{-1}
b_{q,\gamma}^{\pi,(j)},
\qquad
0\leq j\leq L.
\]
Then
\[
V_{q,\gamma}^{\pi,(j)}
=
(I-\gamma P_\pi)^{-1}
b_{q,\gamma}^{\pi,(j)}
\]
for every $0\leq j\leq L$, and the resolvent identity gives
\begin{align}
\widehat V_{q,\gamma}^{\pi,(j)}
-
V_{q,\gamma}^{\pi,(j)}
=
\gamma
(I-\gamma\widehat P_\pi)^{-1}
(\widehat P_\pi-P_\pi)
V_{q,\gamma}^{\pi,(j)}.
\label{eq:hierarchy-resolvent}
\end{align}

Once \cref{eq:li-hierarchy-condition} holds, the hierarchy recursion
is deterministic and does not require $\pi$ to be independent of
$\widehat P$. Since
\[
L
=
\left\lceil
\log\frac{e}{1-\gamma}
\right\rceil,
\]
the condition is available for all hierarchy levels required by
\citet[Lemma~2]{li2020breaking}.

By choosing the universal constant $C$ sufficiently large, the
assumption
\[
n
\geq
\frac{C\ell}{1-\gamma}
\]
implies
\[
n
\geq
\frac{16e^2\beta_1}{1-\gamma}.
\]
Therefore,
\citet[Lemma~2, Eq.~(80)]{li2020breaking} yields
\begin{align}
\left\|
\widehat V_{q,\gamma}^{\pi}
-
V_{q,\gamma}^{\pi}
\right\|_\infty
\leq&
\frac{\gamma\beta_1}
{n(1-\gamma)}
\left\|
V_{q,\gamma}^{\pi}
\right\|_\infty
\nonumber\\
&+
4\gamma
\sqrt{\frac{\beta_1}{n}}\,
\left\|
(I-\gamma P_\pi)^{-1}
\sqrt{
\operatorname{Var}_{P_\pi}
\bigl(V_{q,\gamma}^{\pi}\bigr)
}
\right\|_\infty.
\label{eq:li-instance-dependent}
\end{align}

Since $q\in[0,1]$,
\[
\left\|
V_{q,\gamma}^{\pi}
\right\|_\infty
\leq
\frac{1}{1-\gamma}.
\]
Also,
\[
\sqrt{
\operatorname{Var}_{P_\pi}
\bigl(V_{q,\gamma}^{\pi}\bigr)
}
=
\sigma_{q,\gamma}^{\pi}.
\]
Substituting these relations and $\beta_1=2\ell$ into
\cref{eq:li-instance-dependent}, and absorbing numerical constants
into a universal $C$, gives
\[
\left\|
V_{q,\gamma}^{\pi}
-
\widehat V_{q,\gamma}^{\pi}
\right\|_\infty
\leq
C\left[
\sqrt{\frac{\ell}{n}}\,
\gamma
\left\|
(I-\gamma P_\pi)^{-1}
\sigma_{q,\gamma}^{\pi}
\right\|_\infty
+
\frac{\ell}{n(1-\gamma)^2}
\right].
\]
This proves \cref{eq:data-dependent-plugin}.
\end{proof}

On the intersection of the gap event in
Lemma~\ref{lemma:discounted-gap} and the event
$\mathcal E_{\mathrm{hier}}$ in
Lemma~\ref{lemma:bernstein-hierarchy}, fix
$\lambda\in\Lambda$ and $q\in\{r,c\}$. For every state $x$, define
\[
z_x
:=
(x,\pihat_\lambda(x)).
\]
Lemma~\ref{lemma:row-candidates} gives
\[
\pihat_\lambda
\in
\Pi_{z_x,\lambda}.
\]
Hence \cref{eq:hierarchy-row-bound} holds at every policy-selected row
$z_x$ for the same policy $\pihat_\lambda$. Stacking these rowwise
inequalities verifies the hypothesis of
Lemma~\ref{lemma:hierarchy-plugin}. Therefore,
\cref{eq:data-dependent-plugin} holds simultaneously for
\[
\pi=\pihat_\lambda,
\qquad
\lambda\in\Lambda,
\qquad
q\in\{r,c\}.
\]

\oracleconcentration*

\begin{proof}
Intersect the gap event of Lemma~\ref{lemma:discounted-gap} and the hierarchy event of Lemma~\ref{lemma:bernstein-hierarchy}. By Lemma~\ref{lemma:row-candidates}, the actual oracle $\pihat_\lambda$ belongs to every rowwise candidate family, so it satisfies the complete hierarchy simultaneously for every $\lambda\in\Lambda$ and both $q\in\{r,c\}$. Lemma~\ref{lemma:hierarchy-plugin} therefore gives
\[
\left\|V_{q,\gamma}^{\pihat_\lambda}-\widehat V_{q,\gamma}^{\pihat_\lambda}\right\|_\infty
\leq
C\left[
\sqrt{\frac\ell n}\,
\gamma\left\|(I-\gamma P_{\pihat_\lambda})^{-1}\sigma_{q,\gamma}^{\pihat_\lambda}\right\|_\infty
+
\frac{\ell}{n(1-\gamma)^2}
\right].
\]
Assumption~\ref{assum:policywise-variance} applies to each realized deterministic oracle policy. Multiplying by $1-\gamma$ gives
\[
(1-\gamma)
\left\|V_{q,\gamma}^{\pihat_\lambda}-\widehat V_{q,\gamma}^{\pihat_\lambda}\right\|_\infty
\leq
C\left[
\sqrt{\frac{(B+H)\ell}{n}}
+
\frac{\ell}{n(1-\gamma)}
\right],
\]
which is at most $\alpha$ under the sample-size condition in Theorem~\ref{thm:cross-discount-conc}. The burn-in condition in Lemma~\ref{lemma:hierarchy-plugin} is implied by the same sample size because $\alpha<1$.

For the fixed $O(1)$ policies, apply Lemma~\ref{lemma:fixed-discounted-conc} with a union bound. If $\mu=\sum_i\theta_i\pi_i$ is an external mixture supported on covered policies, then linearity and the triangle inequality give
\[
(1-\gamma)
\left|
V_{q,\gamma}^{\mu}(s)-\widehat V_{q,\gamma}^{\mu}(s)
\right|
\leq
\sum_i\theta_i(1-\gamma)
\left\|V_{q,\gamma}^{\pi_i}-\widehat V_{q,\gamma}^{\pi_i}\right\|_\infty
\leq\alpha.
\]
\end{proof}

\newpage
\section{Supporting Lemmas for the Upper Bound}
\label{app:proofs-ad}

\begin{thmbox}
\begin{lemma}[Bounding the discounted sensitivity error]
\label{lemma:sensitivity}
Let \(b_1>b_0\), and let
\[
\pi_1
\in
\argmax_{\pi\in\Pi_{\mathrm{mix}}}
(1-\gamma)\widehat V_{r_p,\gamma}^{\pi}(s)
\quad
\text{s.t.}
\quad
(1-\gamma)\widehat V_{c,\gamma}^{\pi}(s)
\geq
b_1,
\]
and
\[
\pi_0
\in
\argmax_{\pi\in\Pi_{\mathrm{mix}}}
(1-\gamma)\widehat V_{r_p,\gamma}^{\pi}(s)
\quad
\text{s.t.}
\quad
(1-\gamma)\widehat V_{c,\gamma}^{\pi}(s)
\geq
b_0.
\]
If \(\lambda_1^*\) is an optimal dual multiplier for the problem with
threshold \(b_1\), then
\[
0
\leq
(1-\gamma)\widehat V_{r_p,\gamma}^{\pi_0}(s)
-
(1-\gamma)\widehat V_{r_p,\gamma}^{\pi_1}(s)
\leq
(b_1-b_0)\lambda_1^*.
\]
\end{lemma}
\end{thmbox}

\begin{proof}
By strong duality for the problem with threshold \(b_1\),
\[
(1-\gamma)\widehat V_{r_p,\gamma}^{\pi_1}(s)
=
\max_{\pi\in\Pi_{\mathrm{mix}}}
\left\{
(1-\gamma)\widehat V_{r_p,\gamma}^{\pi}(s)
+
\lambda_1^*
\left(
(1-\gamma)\widehat V_{c,\gamma}^{\pi}(s)-b_1
\right)
\right\}.
\]
Evaluating the maximum at \(\pi_0\) gives
\begin{align*}
(1-\gamma)\widehat V_{r_p,\gamma}^{\pi_1}(s)
&\geq
(1-\gamma)\widehat V_{r_p,\gamma}^{\pi_0}(s)
+
\lambda_1^*
\left(
(1-\gamma)\widehat V_{c,\gamma}^{\pi_0}(s)-b_1
\right)
\\
&\geq
(1-\gamma)\widehat V_{r_p,\gamma}^{\pi_0}(s)
-
(b_1-b_0)\lambda_1^*.
\end{align*}
Since the problem with threshold \(b_0\) is less constrained,
\[
(1-\gamma)\widehat V_{r_p,\gamma}^{\pi_0}(s)
\geq
(1-\gamma)\widehat V_{r_p,\gamma}^{\pi_1}(s).
\]
Combining the two inequalities proves the result.
\end{proof}

\begin{thmbox}
\begin{lemma}[Bias span of normalized Lagrangian rewards]
\label{lemma:lagrangian-span}
For \(\lambda\geq0\), define
\[
q_\lambda
:=
\frac{
r+\lambda c
}{
1+\lambda
}.
\]
For every \(\pi\in\Pi_{\mathrm{det}}\),
\[
\spannorm{
h_{q_\lambda}^{\pi}
}
\leq
H.
\]
Equivalently,
\[
\spannorm{
h_{r+\lambda c}^{\pi}
}
\leq
(1+\lambda)H.
\]
The transient-time parameter remains \(B\).
\end{lemma}
\end{thmbox}

\begin{proof}
By linearity of the gain and bias in the one-step signal,
\[
h_{q_\lambda}^{\pi}
=
\frac{
h_r^\pi+\lambda h_c^\pi
}{
1+\lambda
}.
\]
Therefore,
\[
\spannorm{
h_{q_\lambda}^{\pi}
}
\leq
\frac{
\spannorm{h_r^\pi}
+
\lambda\spannorm{h_c^\pi}
}{
1+\lambda
}
\leq
\frac{
H+\lambda H
}{
1+\lambda
}
=
H.
\]
The recurrent and transient states depend only on the transition kernel
and the policy, so changing the signal does not change \(B\).
\end{proof}

\section{Proof of Lower Bound}
\label{app:proofs-lb}

\lbgen*

\begin{figure}[t]
\centering
\IfFileExists{gmdp.png}{\includegraphics[width=0.72\textwidth]{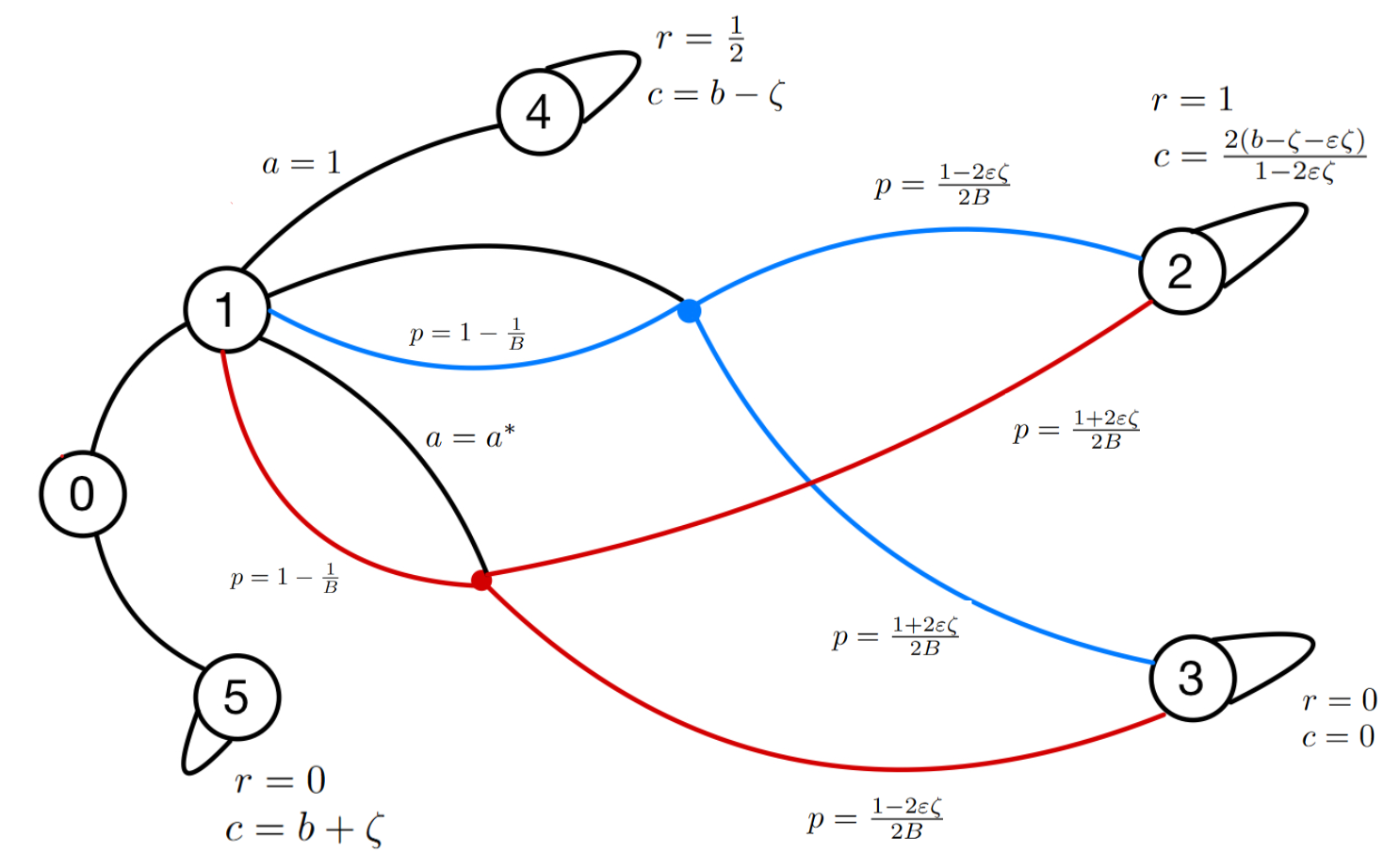}}{\fbox{\parbox[c][4.0cm][c]{0.68\textwidth}{\centering Place \texttt{gmdp.png} here.}}}
\caption{The six-state component used in the transient hard family.}
\label{fig:transient-component}
\end{figure}

\subsection{Hard family}

The active state space is
\[
\{0,1,2,3,4,5\}.
\]
State \(0\) is the initial state. The remaining \(S-6\) states are
unreachable dummy absorbing states. Every dummy state is absorbing
under every action and has zero reward and zero constraint reward.
Every state has \(A\) actions.

At state \(0\), action \(a_1\) leads deterministically to state \(5\),
while every action \(a\neq a_1\) leads deterministically to state \(1\).

At state \(1\), action \(a_1\) leads deterministically to state \(4\).
For every \(a\neq a_1\), the base transition law is
\[
Q_0
=
\operatorname{Cat}
\left(
1-\frac1B,
\frac{1-2\epsilon\zeta}{2B},
\frac{1+2\epsilon\zeta}{2B}
\right)
\]
over next states \(1,2,3\). In the perturbed instance \(M_{a^*}\), only
action \(a^*\) is changed, with transition law
\[
Q_1
=
\operatorname{Cat}
\left(
1-\frac1B,
\frac{1+2\epsilon\zeta}{2B},
\frac{1-2\epsilon\zeta}{2B}
\right).
\]

States \(2,3,4,5\) are absorbing under every action. Their reward values
are
\[
r(2)=1,
\qquad
r(3)=0,
\qquad
r(4)=\frac12,
\qquad
r(5)=0.
\]
Their constraint rewards are
\[
c(2)
=
\frac{
2(b-\zeta-\epsilon\zeta)
}{
1-2\epsilon\zeta
},
\qquad
c(3)=0,
\]
\[
c(4)
=
b-\zeta,
\qquad
c(5)
=
b+\zeta.
\]
All transient-state rewards and constraint rewards are zero.

Set
\[
b
=
\frac12.
\]
Assume
\[
0<\epsilon\leq\frac1{16},
\qquad
0<\zeta\leq\frac14.
\]
These assumptions ensure that all transition probabilities and all
reward and constraint values are valid. In particular,
\[
b-\zeta-\epsilon\zeta
\geq
\frac12-\frac14-\frac1{64}
>
0,
\]
and
\[
0
\leq
c(2)
\leq
1.
\]

The actual transient-time parameter of the instance is between \(B\)
and \(B+1\): from state \(1\), a non-\(a_1\) action remains at state
\(1\) for an expected \(B\) visits, while state \(0\) adds at most one
deterministic entry step. Thus the actual parameter is
\(\Theta(B)\).

\begin{thmbox}
\begin{lemma}[Occupancy representation of external mixtures]
\label{lemma:external-mixture-occupancy}
For an external mixture
\[
\mu
=
\sum_i
\theta_i\pi_i,
\]
let:

\begin{itemize}
\item \(\mu_0\) be the total mixture weight assigned to deterministic
policies choosing \(a_1\) at state \(0\);

\item \(\mu_1\) be the total mixture weight assigned to deterministic
policies choosing \(a\neq a_1\) at state \(0\) and \(a_1\) at state
\(1\);

\item \(\mu_2\) be the total mixture weight assigned to deterministic
policies choosing \(a\neq a_1\) at state \(0\) and an action
\(a\neq a_1\) at state \(1\).
\end{itemize}

Then
\[
\mu_0+\mu_1+\mu_2=1,
\qquad
\mu_0,\mu_1,\mu_2\geq0.
\]

In the base instance,
\[
\rho_r^\mu(0)
=
\frac12\mu_1
+
\left(
\frac12-\epsilon\zeta
\right)\mu_2
\]
and
\[
\rho_c^\mu(0)
=
(b+\zeta)\mu_0
+
(b-\zeta)\mu_1
+
(b-\zeta-\epsilon\zeta)\mu_2.
\]

In the perturbed instance, let \(\mu_3\) be the total mixture weight
assigned to deterministic policies choosing \(a\neq a_1\) at state
\(0\) and choosing \(a^*\) at state \(1\). Set
\[
\mu_2^c
:=
\mu_2-\mu_3.
\]
Then the reward and constraint values equal the objective and
constraint expressions in LP3 and LP4 below.

Conversely, every optimizer used in the LP calculations below is
attainable by an external mixture of stationary deterministic
policies.
\end{lemma}
\end{thmbox}

\begin{proof}
Every deterministic policy belongs to exactly one of the three base
modes, according to its action at states \(0\) and \(1\). Averaging the
mode indicators over the external-mixture weights gives nonnegative
numbers summing to one.

If the deterministic policy chooses \(a_1\) at state \(0\), it reaches
state \(5\) and obtains reward \(0\) and constraint value \(b+\zeta\).

If it reaches state \(1\) and chooses \(a_1\), it reaches state \(4\)
and obtains reward \(1/2\) and constraint value \(b-\zeta\).

If it chooses a non-\(a_1\) action at state \(1\) under the base
instance, the conditional probability of eventual absorption in state
\(2\) is
\[
\frac{1-2\epsilon\zeta}{2},
\]
while the probability of eventual absorption in state \(3\) is
\[
\frac{1+2\epsilon\zeta}{2}.
\]
Therefore, its average reward is
\[
\frac{1-2\epsilon\zeta}{2}
=
\frac12-\epsilon\zeta.
\]
Its average constraint value is
\[
\frac{1-2\epsilon\zeta}{2}
\cdot
\frac{
2(b-\zeta-\epsilon\zeta)
}{
1-2\epsilon\zeta
}
=
b-\zeta-\epsilon\zeta.
\]

The perturbed formulas follow by separating the \(a^*\)-mode from the
other non-\(a_1\) modes.

Finally, the construction contains deterministic policies
corresponding to every mode used by the LP optimizers. External
randomization among these deterministic policies realizes each
optimizer appearing below.
\end{proof}

\begin{thmbox}
\begin{lemma}[Slater constant of the transient hard family]
\label{lemma:hard-slater}
The Slater constant of every instance in the transient hard family is
exactly \(\zeta\).
\end{lemma}
\end{thmbox}

\begin{proof}
The safe deterministic policy chooses \(a_1\) at state \(0\), reaches
state \(5\), and obtains constraint value
\[
b+\zeta.
\]
Hence the Slater constant is at least \(\zeta\).

The remaining policy-level constraint values are
\[
b-\zeta,
\qquad
b-\zeta-\epsilon\zeta,
\qquad
c_\star,
\]
where the perturbed \(a^*\)-mode has value
\[
c_\star
:=
\frac{
(b-\zeta-\epsilon\zeta)(1+2\epsilon\zeta)
}{
1-2\epsilon\zeta
}.
\]
For \(b=1/2\),
\begin{align*}
c_\star-(b+\zeta)
&=
\frac{
\zeta(\epsilon-2)
-
2\epsilon^2\zeta^2
}{
1-2\epsilon\zeta
}
\\
&\leq
0.
\end{align*}
Thus every deterministic policy has constraint value at most
\(b+\zeta\). Since an external mixture has a constraint value equal to
a convex combination of deterministic-policy values, it cannot exceed
\(b+\zeta\). Therefore, the Slater constant is exactly \(\zeta\).
\end{proof}

\subsection{Base-instance separation}

By Lemma~\ref{lemma:external-mixture-occupancy}, the base optimization
problem is
\begin{align}
\max\quad
&
\frac12\mu_1
+
\left(
\frac12-\epsilon\zeta
\right)\mu_2
\tag{LP1}
\label{lp1}
\\
\text{s.t.}\quad
&
\mu_0+\mu_1+\mu_2=1,
\nonumber
\\
&
(b+\zeta)\mu_0
+
(b-\zeta)\mu_1
+
(b-\zeta-\epsilon\zeta)\mu_2
\geq
b,
\nonumber
\\
&
\mu_0,\mu_1,\mu_2\geq0.
\nonumber
\end{align}

Since \(b=1/2\), eliminating \(\mu_0\) shows that the constraint is
equivalent to
\[
2\mu_1+(2+\epsilon)\mu_2
\leq
1.
\]
The reward per unit of the resource constraint is
\[
\frac{1/2}{2}
=
\frac14
\]
for \(\mu_1\), whereas
\[
\frac{
1/2-\epsilon\zeta
}{
2+\epsilon
}
<
\frac14
\]
for \(\mu_2\). Thus the unique optimum is
\[
\mu_0
=
\frac12,
\qquad
\mu_1
=
\frac12,
\qquad
\mu_2
=
0,
\]
with value
\[
\rho_{M_0}^*(0)
=
\frac14.
\]

Define the normalized action occupancy on the entire output space by
\[
\mu_1'
:=
\begin{cases}
\mu_1/(1-\mu_0),&\mu_0<1,\\
0,&\mu_0=1.
\end{cases}
\]
When $\mu_0=1$, the reward is zero, so this convention does not affect the separation of successful outputs and makes the testing event measurable for every possible algorithm output.

Suppose
\[
\mu_1'
\leq
\frac23.
\]
The corresponding closed LP is
\begin{align}
\max\quad
&
\frac12\mu_1
+
\left(
\frac12-\epsilon\zeta
\right)\mu_2
\tag{LP2}
\label{lp2}
\\
\text{s.t.}\quad
&
\mu_0+\mu_1+\mu_2=1,
\nonumber
\\
&
3\mu_1
\leq
2(1-\mu_0),
\nonumber
\\
&
(b+\zeta)\mu_0
+
(b-\zeta)\mu_1
+
(b-\zeta-\epsilon\zeta)\mu_2
\geq
b,
\nonumber
\\
&
\mu_0,\mu_1,\mu_2\geq0.
\nonumber
\end{align}

The normalized-occupancy constraint is equivalent to
\[
\mu_1
\leq
2\mu_2.
\]
At the optimum, both
\[
2\mu_1+(2+\epsilon)\mu_2
\leq
1
\]
and
\[
\mu_1\leq2\mu_2
\]
are active. Therefore,
\[
\mu_2
=
\frac1{6+\epsilon},
\qquad
\mu_1
=
\frac2{6+\epsilon},
\qquad
\mu_0
=
\frac{3+\epsilon}{6+\epsilon}.
\]
The optimal value of LP2 is
\[
\operatorname{val}(\mathrm{LP2})
=
\frac{
\frac32-\epsilon\zeta
}{
6+\epsilon
}.
\]
The exact optimality gap is
\begin{align*}
\rho_{M_0}^*(0)
-
\operatorname{val}(\mathrm{LP2})
&=
\frac14
-
\frac{
\frac32-\epsilon\zeta
}{
6+\epsilon
}
\\
&=
\frac{
\epsilon(1+4\zeta)
}{
4(6+\epsilon)
}.
\end{align*}
For \(\epsilon\leq1/16\),
\[
\frac{
\epsilon(1+4\zeta)
}{
4(6+\epsilon)
}
\geq
\frac{4\epsilon}{97}
>
\frac{\epsilon}{48}.
\]
Therefore, every \(\epsilon/48\)-optimal strictly feasible policy under
\(M_0\) satisfies
\[
\mu_1'
>
\frac23.
\]

\subsection{Perturbed-instance separation}

In the perturbed instance, define \(\mu_3\) as in
Lemma~\ref{lemma:external-mixture-occupancy}, and set
\[
\mu_2^c
=
\mu_2-\mu_3.
\]
The optimization problem is
\begin{align}
\max\quad
&
\frac12\mu_1
+
\left(
\frac12-\epsilon\zeta
\right)\mu_2^c
+
\left(
\frac12+\epsilon\zeta
\right)\mu_3
\tag{LP3}
\label{lp3}
\\
\text{s.t.}\quad
&
\mu_0+\mu_1+\mu_2^c+\mu_3=1,
\nonumber
\\
&
(b+\zeta)\mu_0
+
(b-\zeta)\mu_1
+
(b-\zeta-\epsilon\zeta)\mu_2^c
+
c_\star\mu_3
\geq
b,
\nonumber
\\
&
\mu_0,\mu_1,\mu_2^c,\mu_3\geq0.
\nonumber
\end{align}

Define
\[
\chi_{\epsilon,\zeta}
:=
\frac{
2-\epsilon+2\epsilon^2\zeta
}{
1-2\epsilon\zeta
}.
\]
Eliminating \(\mu_0\), the constraint in LP3 becomes
\[
2\mu_1
+
(2+\epsilon)\mu_2^c
+
\chi_{\epsilon,\zeta}\mu_3
\leq
1.
\]

For
\[
\epsilon\leq\frac1{16},
\qquad
\zeta\leq\frac14,
\]
we have
\[
\frac{
1/2-\epsilon\zeta
}{
2+\epsilon
}
<
\frac14
<
\frac{
1/2+\epsilon\zeta
}{
\chi_{\epsilon,\zeta}
}.
\]
Hence the unique optimum of LP3 uses only the safe mode and the
perturbed \(a^*\)-mode:
\[
\mu_1
=
\mu_2^c
=
0,
\qquad
\mu_3
=
\frac1{\chi_{\epsilon,\zeta}},
\qquad
\mu_0
=
1-\frac1{\chi_{\epsilon,\zeta}}.
\]
The exact optimal value is
\[
\rho_{M_{a^*}}^*(0)
=
\frac{
\frac12+\epsilon\zeta
}{
\chi_{\epsilon,\zeta}
}.
\]

Suppose instead that
\[
\mu_1'
\geq
\frac23.
\]
The corresponding closed LP is
\begin{align}
\max\quad
&
\frac12\mu_1
+
\left(
\frac12-\epsilon\zeta
\right)\mu_2^c
+
\left(
\frac12+\epsilon\zeta
\right)\mu_3
\tag{LP4}
\label{lp4}
\\
\text{s.t.}\quad
&
\mu_0+\mu_1+\mu_2^c+\mu_3=1,
\nonumber
\\
&
3\mu_1
\geq
2(1-\mu_0),
\nonumber
\\
&
(b+\zeta)\mu_0
+
(b-\zeta)\mu_1
+
(b-\zeta-\epsilon\zeta)\mu_2^c
+
c_\star\mu_3
\geq
b,
\nonumber
\\
&
\mu_0,\mu_1,\mu_2^c,\mu_3\geq0.
\nonumber
\end{align}

The normalized-occupancy constraint is equivalent to
\[
\mu_1
\geq
2(\mu_2^c+\mu_3).
\]
Write
\[
u
:=
\mu_1-2(\mu_2^c+\mu_3)
\geq
0.
\]
After eliminating \(\mu_0\), the resource constraint becomes
\[
2u
+
(6+\epsilon)\mu_2^c
+
(4+\chi_{\epsilon,\zeta})\mu_3
\leq
1,
\]
and the objective becomes
\[
\frac12u
+
\left(
\frac32-\epsilon\zeta
\right)\mu_2^c
+
\left(
\frac32+\epsilon\zeta
\right)\mu_3.
\]

Under the stated parameter range,
\[
\frac{
\frac32-\epsilon\zeta
}{
6+\epsilon
}
<
\frac14
<
\frac{
\frac32+\epsilon\zeta
}{
4+\chi_{\epsilon,\zeta}
}.
\]
Thus, the optimum of LP4 is
\[
u
=
\mu_2^c
=
0,
\qquad
\mu_3
=
\frac1{4+\chi_{\epsilon,\zeta}},
\qquad
\mu_1
=
\frac2{4+\chi_{\epsilon,\zeta}}.
\]
Therefore,
\[
\operatorname{val}(\mathrm{LP4})
=
\frac{
\frac32+\epsilon\zeta
}{
4+\chi_{\epsilon,\zeta}
}.
\]

The exact gap from the perturbed optimum is
\begin{align*}
&
\rho_{M_{a^*}}^*(0)
-
\operatorname{val}(\mathrm{LP4})
\\
&=
\frac{
\frac12+\epsilon\zeta
}{
\chi_{\epsilon,\zeta}
}
-
\frac{
\frac32+\epsilon\zeta
}{
4+\chi_{\epsilon,\zeta}
}
\\
&=
\frac{
2+4\epsilon\zeta-\chi_{\epsilon,\zeta}
}{
\chi_{\epsilon,\zeta}
(4+\chi_{\epsilon,\zeta})
}
\\
&=
\frac{
\epsilon
\left[
1-2\epsilon\zeta(1+4\zeta)
\right]
}{
(1-2\epsilon\zeta)
\chi_{\epsilon,\zeta}
(4+\chi_{\epsilon,\zeta})
}.
\end{align*}

For
\[
\epsilon\leq\frac1{16},
\qquad
\zeta\leq\frac14,
\]
we have
\[
1-2\epsilon\zeta(1+4\zeta)
\geq
\frac{15}{16}
\]
and
\[
\chi_{\epsilon,\zeta}
\leq
\frac{64}{31}.
\]
Moreover,
\[
(1-2\epsilon\zeta)
\chi_{\epsilon,\zeta}
(4+\chi_{\epsilon,\zeta})
\leq
\frac{64}{31}
\left(
4+\frac{64}{31}
\right)
<
13.
\]
Consequently,
\[
\rho_{M_{a^*}}^*(0)
-
\operatorname{val}(\mathrm{LP4})
\geq
\frac{15\epsilon}{16\cdot13}
>
\frac{\epsilon}{48}.
\]
Thus, every \(\epsilon/48\)-optimal strictly feasible policy under
\(M_{a^*}\) satisfies
\[
\mu_1'
<
\frac23.
\]

\subsection{Testing reduction and KL divergence}

The statistic \(\mu_1'\) is determined solely by the output
external-mixture weights and by the actions selected by its
deterministic policies at states \(0\) and \(1\). Hence
\[
E
:=
\left\{
\mu_1'<\frac23
\right\}
\]
is a common measurable event under the base and perturbed transcript
distributions.

If an algorithm succeeds with probability at least \(3/4\) on both
instances, then
\[
\Pr_{M_0}(E)
\leq
\frac14
\]
and
\[
\Pr_{M_{a^*}}(E)
\geq
\frac34.
\]

For
\[
\epsilon\zeta
\leq
\frac14,
\]
the distinguished-row KL divergence satisfies
\begin{align*}
D_{\mathrm{KL}}(Q_1\|Q_0)
&=
\frac{
1+2\epsilon\zeta
}{
2B
}
\log
\frac{
1+2\epsilon\zeta
}{
1-2\epsilon\zeta
}
\\
&\quad
+
\frac{
1-2\epsilon\zeta
}{
2B
}
\log
\frac{
1-2\epsilon\zeta
}{
1+2\epsilon\zeta
}
\\
&=
\frac{
2\epsilon\zeta
}{
B
}
\log
\frac{
1+2\epsilon\zeta
}{
1-2\epsilon\zeta
}
\\
&\leq
\frac{
16\epsilon^2\zeta^2
}{
B(1-4\epsilon^2\zeta^2)
}
\\
&\leq
\frac{
32\epsilon^2\zeta^2
}{
B
}.
\end{align*}

Under balanced sampling, the algorithm obtains \(n\) independent
samples from the distinguished row. Since the two instances differ
only at this row,
\[
D_{\mathrm{KL}}
\left(
\mathbb P_{M_{a^*}}
\|
\mathbb P_{M_0}
\right)
=
n
D_{\mathrm{KL}}(Q_1\|Q_0).
\]
Binary data processing yields
\[
n
D_{\mathrm{KL}}(Q_1\|Q_0)
\geq
\operatorname{kl}
\left(
\frac34,
\frac14
\right)
=:c_{\mathrm{test}}>0.
\]
Therefore,
\[
n
\geq
c
\frac{
B
}{
\epsilon^2\zeta^2
}
\]
for a universal constant \(c>0\). Since balanced sampling uses
\(SA\,n\) samples,
\[
N_{\mathrm{tot}}
\geq
c
\frac{
SAB
}{
\epsilon^2\zeta^2
}.
\]

\subsection{Transient-time and span calibration}

The expected time from state \(1\) to absorption under a non-\(a_1\)
action is \(B\). From state \(0\), one additional deterministic step
may be required. Hence the actual uniform transient-time parameter
\(B_{\mathrm{inst}}\) satisfies
\[
B
\leq
B_{\mathrm{inst}}
\leq
B+1
\leq
2B.
\]

Normalize the bias to be zero on every absorbing recurrent state. For
a transient state \(x\),
\begin{align*}
|h_q^\pi(x)|
&\leq
\E_x^\pi
\left[
\sum_{t=0}^{T_{\cR^\pi}-1}
\left|
q(S_t,A_t)-\rho_q^\pi(S_t)
\right|
\right]
\\
&\leq
\E_x^\pi[T_{\cR^\pi}]
\\
&\leq
B+1.
\end{align*}
Therefore,
\[
\spannorm{h_q^\pi}
\leq
2(B+1)
\leq
4B
\]
for \(q\in\{r,c\}\), and hence
\[
H
\leq
4B.
\]
Consequently,
\[
B+H
=
\Theta(B)
\]
on the transient-dominated hard family.

\begin{thmbox}
\begin{lemma}[Variance condition for the transient hard family]
\label{lemma:hard-family-variance}
For every deterministic policy \(\pi\), every \(q\in\{r,c\}\), and
every \(\gamma\in(0,1)\) satisfying
\[
B
\leq
\frac1{1-\gamma},
\]
the transient hard family satisfies
\[
\gamma
\left\|
(I-\gamma P_\pi)^{-1}
\sigma_{q,\gamma}^{\pi}
\right\|_\infty
\leq
C
\frac{
\sqrt B
}{
1-\gamma
}
\]
for a universal constant \(C>0\).
\end{lemma}
\end{thmbox}

\begin{proof}
Every absorbing state has zero one-step conditional variance. If the
policy selects \(a_1\) at state \(1\), the transition is deterministic
and the variance is also zero.

Suppose the policy selects \(a\neq a_1\). Since
\[
0
\leq
V_{q,\gamma}^{\pi}
\leq
\frac1{1-\gamma},
\]
the range of \(V_{q,\gamma}^{\pi}\) is at most
\(1/(1-\gamma)\). The total exit probability from state \(1\) is
\(1/B\). Therefore,
\[
\operatorname{Var}_{P_\pi}
\left(
V_{q,\gamma}^{\pi}
\right)(1)
\leq
\frac{
1
}{
B(1-\gamma)^2
}.
\]
Thus,
\[
\sigma_{q,\gamma}^{\pi}(1)
\leq
\frac{
1
}{
\sqrt B(1-\gamma)
}.
\]

The discounted expected number of visits to state \(1\) is
\[
\sum_{t=0}^{\infty}
\left(
\gamma\left(1-\frac1B\right)
\right)^t
=
\frac{
1
}{
1-\gamma(1-1/B)
}
\leq
B.
\]
Therefore,
\[
\gamma
\left[
(I-\gamma P_\pi)^{-1}
\sigma_{q,\gamma}^{\pi}
\right](1)
\leq
\frac{
\sqrt B
}{
1-\gamma
}.
\]

At state \(0\), the transition is deterministic. If it reaches state
\(1\), it does so in one step. Hence the same bound applies at state
\(0\), up to a universal constant. All other states have zero
variance, proving the result.
\end{proof}

The preceding construction proves the transient lower bound
$\widetilde\Omega(SAB/(\epsilon^2\zeta^2))$. Combining it with
Theorem~\ref{thm:wcmdp}, which also applies to balanced algorithms,
and using $\max\{B,H\}\ge(B+H)/2$ completes the proof of
Theorem~
~\ref{thm:general-lower-bound}.

\section{Proof of the Weakly Communicating Lower Bound}
\label{app:proofs-wclb}

\lbcomm*

\begin{figure}[t]
\centering
\begin{minipage}[t]{0.53\textwidth}
\centering
\IfFileExists{wcmdp.png}{\includegraphics[width=\textwidth]{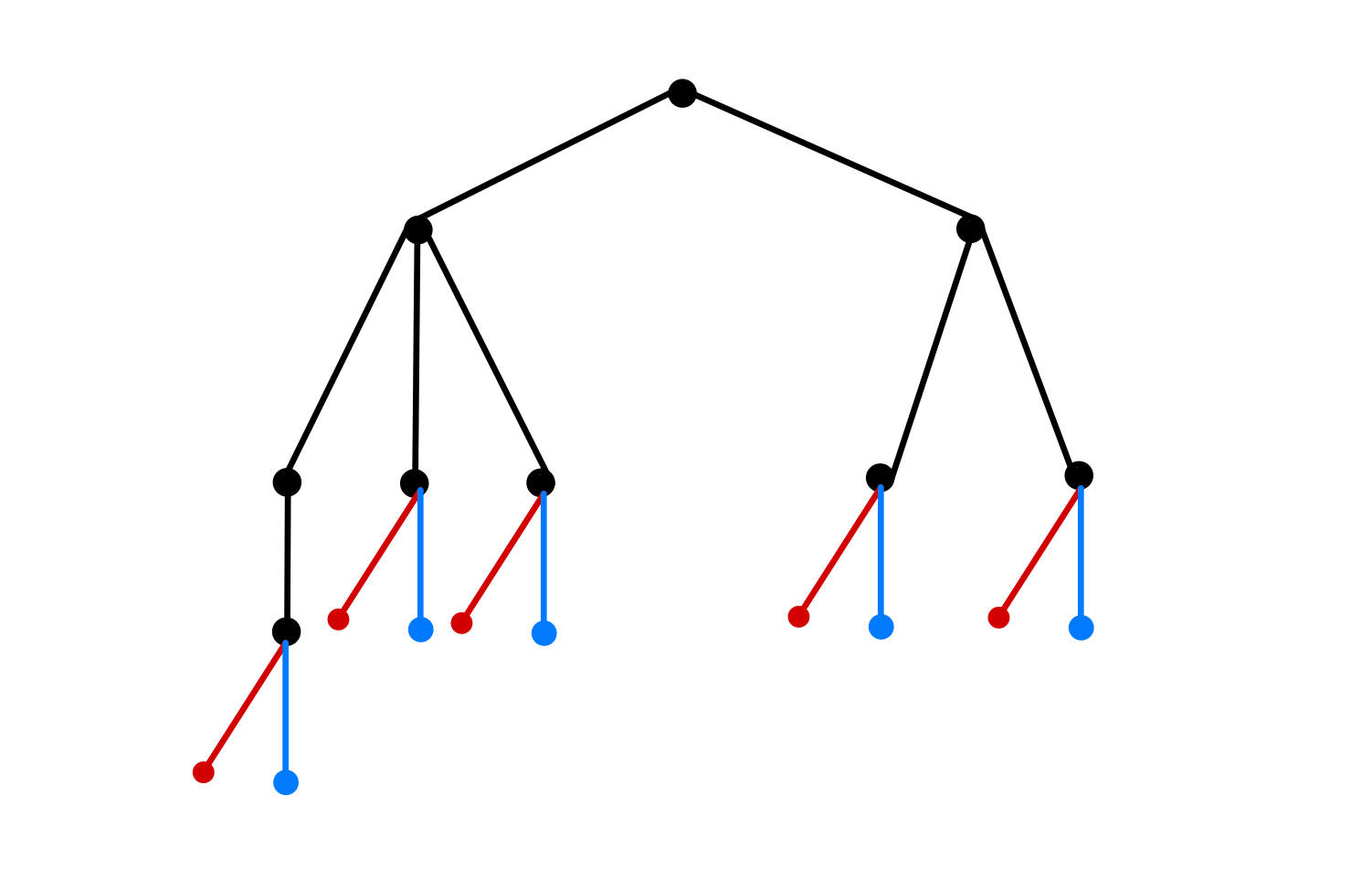}}{\fbox{\parbox[c][3.0cm][c]{0.92\textwidth}{\centering Place \texttt{wcmdp.png} here.}}}
\end{minipage}\hfill
\begin{minipage}[t]{0.41\textwidth}
\centering
\IfFileExists{Component.png}{\includegraphics[width=\textwidth]{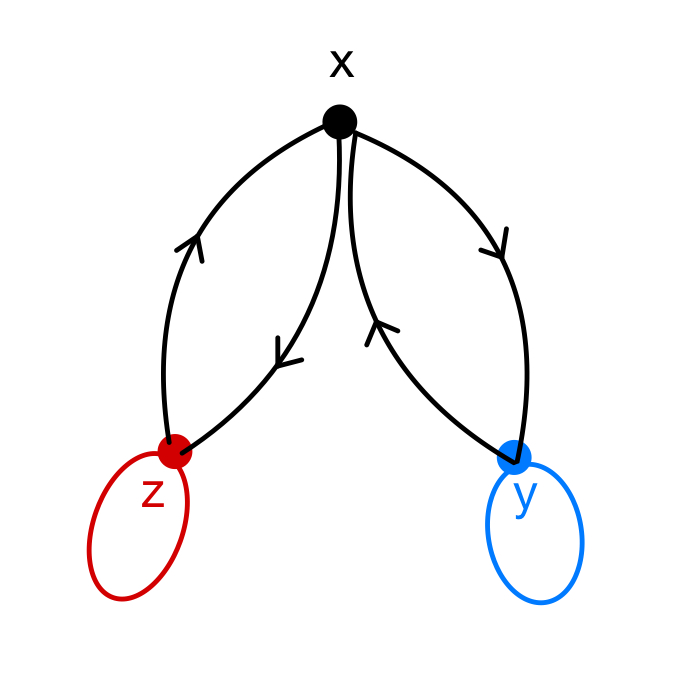}}{\fbox{\parbox[c][3.0cm][c]{0.90\textwidth}{\centering Place \texttt{Component.png} here.}}}
\end{minipage}
\caption{The weakly communicating hard family and its primitive three-state component.}
\label{fig:weak-appendix}
\end{figure}

\subsection{State space and routing structure}

Let $A'=A-1$ and $K=\lfloor(S-1)/4\rfloor$. Construct an $A'$-ary rooted tree with $K$ leaves, at most $2K-1$ vertices, and depth $L\le\lceil\log_{A'}K\rceil+1$. The initial distribution is the point mass at the root, $s=\delta_{x_{\rm root}}$. Each leaf is the entrance state $x_k$ of a component $(x_k,y_k,z_k)$, adding two states per leaf. One additional state $z_{\rm safe}$ forms the safe class. Thus the construction uses at most $4K\le S-1$ states. Any unused states are inserted by subdividing deterministic zero-signal routing edges; all their actions follow the same deterministic routing rule or self-loop with zero signals.

At each internal tree node, at most $A'$ actions move deterministically to its children and one action moves to its parent; unused actions are zero-signal self-loops. At the root, the parent action moves to $z_{\rm safe}$. At $x_k$, action $a_A$ moves deterministically to $z_k$, and every action at $z_k$ moves deterministically to the parent tree state; these transitions have reward and constraint signal zero. At $y_k$, every action moves to $x_k$ with probability $q_0$ and self-loops otherwise. At $z_{\rm safe}$, one action is the safe self-loop with reward $0$ and constraint signal $b+\zeta$; every other action returns deterministically to the root with both signals zero. Consequently the MDP is communicating, hence weakly communicating.

Set $q_0=1/D$, $d=2\epsilon\zeta$, and assume $d\le1/8$. At $x_k$, action $a_1$ moves to $y_k$ with probability $p_0=q_0$ and self-loops otherwise. In the reference instance $M_0$, each action $a_l$, $2\le l\le A'$, moves to $y_k$ with probability
\begin{align}
p_-:=q_0\frac{1+2d}{1-2d}.
\label{eq:p-minus}
\end{align}
Instance $M_{k,l}$ differs only at row $(x_k,a_l)$, where the probability becomes
\begin{align}
p_+:=q_0\frac{1-2d}{1+2d}.
\label{eq:p-plus}
\end{align}
The first $A'$ actions have reward and constraint signal $1$ at $x_k$; all signals at $y_k,z_k$ and on navigation transitions are zero. Set
\begin{align}
b:=\frac12+\zeta,
\qquad
0<\zeta\le\frac18,
\label{eq:weak-threshold}
\end{align}
so the safe constraint value is $b+\zeta=1/2+2\zeta\le1$.

\subsection{Gain pairs, Slater margin, and occupancy representation}

\begin{thmbox}
\begin{lemma}[Component gains]
\label{lemma:weak-component-gains}
If a deterministic policy closes component $k$ using an action whose $x_k\to y_k$ probability is $p$, then its reward and constraint gains are both $g(p)=q_0/(p+q_0)$. Hence
\[
g(p_0)=\frac12,
\qquad
g(p_-)=\frac12-d,
\qquad
g(p_+)=\frac12+d.
\]
\end{lemma}
\end{thmbox}

\begin{proof}
The state $z_k$ is not entered under a slow action. The invariant distribution of the closed two-state class $(x_k,y_k)$ places mass $q_0/(p+q_0)$ on $x_k$, and both signals are the indicator of $x_k$.
\end{proof}

\begin{thmbox}
\begin{lemma}[Exact Slater margin]
\label{lemma:weak-slater}
Every instance has Slater margin exactly $\zeta$.
\end{lemma}
\end{thmbox}

\begin{proof}
The safe policy has constraint value $b+\zeta$. Every component gain is at most $1/2+d$, and $d=2\epsilon\zeta\le\zeta/4$ when $\epsilon\le1/8$. All routing recurrent classes have zero constraint signal. Thus no deterministic policy, and hence no external mixture, has constraint value exceeding $b+\zeta$.
\end{proof}

\begin{thmbox}
\begin{lemma}[External-mixture class representation]
\label{lemma:weak-occupancy}
Fix $M_{k,l}$ and call $(k,l)$ special. Every external mixture can be represented by nonnegative weights $\mu_0,\mu_1,\mu_2,\mu_3,\mu_4$ summing to one, where $\mu_0$ is the safe weight, $\mu_3$ is the special component--action weight, $\mu_1$ is the total baseline-component weight, $\mu_2$ is the total ordinary suboptimal-component weight, and $\mu_4$ is the weight on zero-signal routing classes. With $h_+=1/2+d$, $h_0=1/2$, $h_-=1/2-d$, and $s_+=1/2+2\zeta$,
\begin{align}
\rho_r^\mu(s)&=h_0\mu_1+h_-\mu_2+h_+\mu_3,
\label{eq:weak-reward-exact}\\
\rho_c^\mu(s)&=s_+\mu_0+h_0\mu_1+h_-\mu_2+h_+\mu_3.
\label{eq:weak-constraint-exact}
\end{align}
Conversely, every weight vector used in the optimization below is realizable by an external mixture of deterministic policies.
\end{lemma}
\end{thmbox}

\begin{proof}
Starting from the root, every deterministic policy eventually enters exactly one closed class: the safe singleton, a component under one chosen slow action, or a zero-signal routing cycle. The gains of these classes are given by Lemma~\ref{lemma:weak-component-gains}; linearity of external mixtures gives~\cref{eq:weak-reward-exact,eq:weak-constraint-exact}. Each listed class is reachable by a deterministic routing policy, so the required mixtures are attainable.
\end{proof}

\subsection{Constrained optimum and output separation}

For a hard gain $h<b$, the best feasible mixture of the safe class and that hard class has hard weight $\zeta/(s_+-h)$ and reward
\begin{align}
F(h):=\frac{\zeta h}{s_+-h}.
\label{eq:F-hard}
\end{align}
The function $F$ is strictly increasing, so the optimum in $M_{k,l}$ equals $F(h_+)$. Define the normalized special weight on the entire output space by
\begin{align}
\nu_{k,l}:=
\begin{cases}
\mu_3/(\mu_1+\mu_2+\mu_3),&\mu_1+\mu_2+\mu_3>0,\\
0,&\mu_1+\mu_2+\mu_3=0.
\end{cases}
\label{eq:nu-special}
\end{align}

\begin{thmbox}
\begin{lemma}[Separation of successful outputs]
\label{lemma:weak-separation}
Every strictly feasible external mixture with reward at least $\rho_r^*(s)-\epsilon/48$ in $M_{k,l}$ satisfies $\nu_{k,l}>2/3$.
\end{lemma}
\end{thmbox}

\begin{proof}
Suppose $\nu_{k,l}\le2/3$. Removing zero-signal weight and replacing each ordinary suboptimal class by a baseline class can only increase reward and constraint. Conditional on choosing a hard class, the resulting gain is therefore at most $2h_+/3+h_0/3=1/2+2d/3$. The best feasible reward under the restriction is at most $F(1/2+2d/3)$. A direct calculation gives
\begin{align*}
F\!\left(\frac12+d\right)-F\!\left(\frac12+\frac{2d}{3}\right)
&=\frac{\zeta(d/3)(2\zeta+1/2)}{(2\zeta-d)(2\zeta-2d/3)}
\ge\frac{d}{24\zeta}
=\frac\epsilon{12}
>\frac\epsilon{48}.
\end{align*}
Thus the policy cannot be $\epsilon/48$-optimal.
\end{proof}

The special component--action classes are disjoint. Since the normalized special weights sum to one over all possible special indices whenever the total hard weight is positive, at most one event $E_{k,l}:=\{\nu_{k,l}>2/3\}$ can occur for any output.

\subsection{Adaptive change of measure}

Let $\mathcal J=\{(k,l):1\le k\le K,\ 2\le l\le A'\}$ and $M=|\mathcal J|=\Theta(SA)$. For an arbitrary adaptive generative-model algorithm, let $N_j$ be the number of queries to row $j$ and let $\mathbb P_0,\mathbb P_j$ denote the transcript laws under $M_0,M_j$.

\begin{thmbox}
\begin{lemma}[Distinguished-row divergence]
\label{lemma:weak-KL}
For $d\le1/8$ and $D$ larger than a universal constant,
\[
\operatorname{kl}(\operatorname{Ber}(p_-)\|\operatorname{Ber}(p_+))
\le C_{\rm KL}\frac{d^2}{D}.
\]
\end{lemma}
\end{thmbox}

\begin{proof}
We have $p_--p_+=8q_0d/(1-4d^2)$. The inequality $\operatorname{kl}(u\|v)\le(u-v)^2/[v(1-v)]$, together with $p_+=\Theta(q_0)$ and $1-p_+\ge1/2$, proves the claim.
\end{proof}

Write $x_j=\mathbb P_0(E_j)$. Since at most one $E_j$ occurs, $\sum_jx_j\le1$, so at least $M/2$ indices satisfy $x_j\le2/M\le1/4$. By Lemma~\ref{lemma:weak-separation}, success on $M_j$ gives $\mathbb P_j(E_j)\ge3/4$. Adaptive chain rule and binary data processing yield, for each such $j$,
\begin{align*}
\E_0[N_j]\operatorname{kl}(\operatorname{Ber}(p_-)\|\operatorname{Ber}(p_+))
&=D_{\rm KL}(\mathbb P_0\|\mathbb P_j),\\
D_{\rm KL}(\mathbb P_0\|\mathbb P_j)
&\ge\operatorname{kl}(x_j,\mathbb P_j(E_j))
\ge\operatorname{kl}\!\left(\frac14,\frac34\right)=:c_{\rm test}>0.
\end{align*}
Summing over at least $M/2$ indices gives
\[
\E_0[N_{\rm tot}]
\ge c\frac{MD}{d^2}
=\Omega\!\left(\frac{SAD}{\epsilon^2\zeta^2}\right).
\]

\subsection{Diameter, transient time, span, and variance}

The routing depth is $L=O(\log_{A'}S)$. Entering or leaving a slow component costs $O(D)$ expected time, while routing costs $O(L)$, so the diameter is $O(D+L)=O(D)$. Under any deterministic policy, a component is either a closed recurrent class or is exited after at most one geometric waiting period of mean $D$; the routing path has length $O(L)$. Hence the uniform transient parameter is $O(D)$.

Normalize the canonical bias to have stationary mean zero within each recurrent class. The hitting-time representation bounds all transient-state biases by $O(D)$. On a closed component, the Poisson equations give
\[
h_q^\pi(x_k)-h_q^\pi(y_k)=\frac{q(x_k)-q(y_k)}{p+q_0},
\]
which is $O(D)$. Conversely, for $p_0=q_0=1/D$ and $q=r$,
\[
h_r^\pi(x_k)-h_r^\pi(y_k)=\frac1{p_0+q_0}=\frac D2.
\]
Thus the actual uniform reward/constraint bias-span parameter satisfies $H=\Theta(D)$, and the $D$ lower bound is equivalently an $H$ lower bound.

\begin{thmbox}
\begin{lemma}[Variance condition on the weak hard family]
\label{lemma:weak-variance}
The weakly communicating hard family satisfies Assumption~\ref{assum:policywise-variance} with $B+H=\Theta(D)$ and a universal constant.
\end{lemma}
\end{thmbox}

\begin{proof}
On a closed component with $p,q_0=\Theta(1/D)$, subtraction of the Bellman equations gives
\[
V_{q,\gamma}^\pi(x_k)-V_{q,\gamma}^\pi(y_k)
=\frac{q(x_k)-q(y_k)}{1-\gamma(1-p-q_0)}=O(D)
\]
whenever $(B+H)\le(1-\gamma)^{-1}$. Hence the one-step standard-deviation vector is $O(\sqrt D)$, and multiplication by the resolvent contributes at most $(1-\gamma)^{-1}$. For a transient $y_k$, let $\tau$ be the entrance time into the recurrent set. The stopping-time decomposition
\begin{align*}
[(I-\gamma P_\pi)^{-1}\sigma](x)
={}&\E_x^\pi\!\left[\sum_{t=0}^{\tau-1}\gamma^t\sigma(S_t)\right]
+\E_x^\pi\!\left[\gamma^\tau[(I-\gamma P_\pi)^{-1}\sigma](S_\tau)\right]
\end{align*}
shows that the pre-entrance contribution is $O(D^{3/2})$ and the post-entrance contribution is $O(\sqrt D/(1-\gamma))$. Since $(1-\gamma)^{-1}\gtrsim D$, both are $O(\sqrt D/(1-\gamma))$. All routing transitions are deterministic, proving the claim.
\end{proof}

\section*{Statement of LLM Usage}
This manuscript used large language models solely to assist with language editing and improving the
clarity of writing. All technical content, analysis, and conclusions were conceived, implemented, and
verified entirely by the authors.

\end{document}